\documentclass{article}

\makeatletter
\def\input@path{{styles/}}
\makeatother
\usepackage[preprint]{colm2026_conference}
\usepackage{fontspec}
\normalfont
\usepackage{microtype}
\usepackage{graphicx}
\graphicspath{{./}{figures/}}
\usepackage{trimclip}
\usepackage{xcolor}
\usepackage{booktabs}
\usepackage{array}
\usepackage{colortbl}
\usepackage{float}
\usepackage{enumitem,titletoc}
\usepackage{wrapfig}
\usepackage{subcaption}
\usepackage{tikz}
\usepackage{tcolorbox}
\usepackage{pgfplots}
\pgfplotsset{compat=1.18}
\usepgfplotslibrary{groupplots}

\usepackage{amsmath,amssymb,amsfonts}
\usepackage{booktabs}
\usepackage{multirow}
\usepackage{colortbl}
\usepackage{graphicx}
\usepackage{xcolor}
\usepackage{xspace}
\usepackage{tikz}
\usetikzlibrary{arrows.meta,positioning,fit,calc}

\newcommand{\SMR}[0]{SMR\xspace} 

\definecolor{abldirect}{HTML}{D1495B}
\definecolor{ablvista}{HTML}{2A9D8F}
\definecolor{ablglobal}{HTML}{4A81AD}
\definecolor{abldirectedge}{HTML}{8F2433}
\definecolor{ablvistaedge}{HTML}{1D6F65}
\definecolor{ablglobaledge}{HTML}{1B3B5F}
\definecolor{abldirectband}{HTML}{FDF0F2}
\definecolor{ablvistaband}{HTML}{EBF7F5}
\definecolor{ablglobalband}{HTML}{EBF3F9}
\definecolor{ablmodelband}{HTML}{F1F2F5}
\definecolor{baselinegray}{HTML}{F0F0F0}
\definecolor{ablationbest}{HTML}{EBF3F9}

\def\ispevekreport{1}

\setlist[itemize]{leftmargin=*,topsep=2pt,itemsep=1pt}
\setlist[enumerate]{leftmargin=*,topsep=2pt,itemsep=1pt}
\makeatletter
\newenvironment{suppTable}{%
  \par\noindent\begin{minipage}{\linewidth}%
  \def\@captype{table}\captionsetup{hypcap=false}\centering
}{\end{minipage}\par\medskip}
\newenvironment{suppFigure}{%
  \par\noindent\begin{minipage}{\linewidth}%
  \def\@captype{figure}\captionsetup{hypcap=false}\centering
}{\end{minipage}\par\medskip}
\makeatother

\usepackage[hypertexnames=false]{hyperref}
\usepackage{url}
\usepackage[capitalize,noabbrev]{cleveref}
\crefname{section}{Sec.}{Secs.}
\Crefname{section}{Sec.}{Secs.}
\crefname{table}{Tab.}{Tabs.}
\crefname{figure}{Fig.}{Figs.}

\makeatletter
\DeclareRobustCommand\onedot{\futurelet\@let@token\@onedot}
\def\@onedot{\ifx\@let@token.\else.\null\fi\xspace}

\makeatother

\newcommand{\maketitlesupplementary}{%
  \startcontents[appendix]%
  \appendix
  \setcounter{page}{1}%
  \renewcommand{\theHsection}{supp.\Alph{section}}%
  \renewcommand{\theHsubsection}{supp.\Alph{section}.\arabic{subsection}}%
  \begin{center}
    \Large\textbf{\reporttitle}\\[0.5em]
    Supplementary Material
  \end{center}
  \vspace{0.5em}%
  \begin{center}
    \large\bfseries Contents
  \end{center}
  \printcontents[appendix]{}{1}{\setcounter{tocdepth}{2}}%
  \clearpage
}

\definecolor{abyss}{HTML}{121D36}
\definecolor{polarnight}{HTML}{1A2947}
\definecolor{nebula}{HTML}{2B3F66}
\definecolor{steeltrail}{HTML}{6D87BD}
\definecolor{skytrail}{HTML}{8FA8D8}
\definecolor{starlight}{HTML}{DFE7F5}
\definecolor{warmstar}{HTML}{E8D9C4}
\definecolor{electricblue}{HTML}{3866FF}
\definecolor{covercream}{HTML}{EEF3FA}
\definecolor{coveraccent}{HTML}{3866FF}
\colorlet{pevekpurple}{skytrail}
\colorlet{bargray}{steeltrail}
\colorlet{barlgray}{starlight}
\definecolor{abldirect}{HTML}{D75B5B}
\definecolor{ablvista}{HTML}{4C9A73}
\definecolor{ablglobal}{HTML}{416FBE}
\definecolor{abldirectedge}{HTML}{A43E3E}
\definecolor{ablvistaedge}{HTML}{2F7152}
\definecolor{ablglobaledge}{HTML}{294D91}
\definecolor{abldirectband}{HTML}{F8E9E9}
\definecolor{ablvistaband}{HTML}{E8F3ED}
\definecolor{ablglobalband}{HTML}{E8EEF9}
\definecolor{ablationbest}{HTML}{E5EDF8}

\definecolor{jrcolor}{RGB}{225, 100, 150}

\let\outfit\sffamily

\newcommand{\reportauthors}{Guojian Zhan, Jiahao Xu, Yahui Liu, Yao Hu, Chuan Mu}

\hypersetup{
  pdftitle={ResVision: Spectrum-Matched Resizing and Context-Preserving Inspection for VLMs},
  pdfauthor={\reportauthors},
  colorlinks=true,
  linkcolor=electricblue,
  citecolor=electricblue,
  urlcolor=coveraccent,
  filecolor=electricblue
}
\setcitestyle{numbers,square,comma,sort&compress}

\newcommand{\reporttitle}{ResVision: Spectrum-Matched Resizing and \\
Context-Preserving Inspection for VLMs}
\title{\reporttitle}
\author{\reportauthors}

\makeatletter
\providecommand\@LN@col[1]{}
\providecommand\@LN[2]{}
\makeatother

\begin{document}
\raggedbottom

\fancyhead{}
\renewcommand{\headrulewidth}{0pt}
\color{abyss}
\thispagestyle{empty}

\vspace*{-0.55in}
\begin{tcolorbox}[
  width=\linewidth,
  colback=covercream,
  colframe=covercream,
  boxrule=0pt,
  arc=14pt,
  outer arc=14pt,
  boxsep=0pt,
  left=20pt,
  right=20pt,
  top=13pt,
  bottom=9pt
]
  {\sffamily\fontsize{21.5}{25.5}\selectfont\bfseries\centering
    \advance\leftskip by -12pt\relax\advance\rightskip by -12pt\relax
    \reporttitle\par}
  \vspace{1.05em}
  {\bfseries\centering AllSpark Team\par}

  \vspace{0.55em}
  \begingroup
  \normalfont\fontsize{9.5}{10.5}\selectfont
  \setlength{\parindent}{0pt}
  \setlength{\parskip}{0pt}
  \renewenvironment{quote}{%
    \list{}{\setlength{\leftmargin}{10pt}\setlength{\rightmargin}{10pt}}%
    \item\relax
  }{\endlist}
  \begin{abstract}
Even on \textbf{information-complete} tasks, where the input image contains all the evidence needed to answer, a vision-language model can benefit from seeing the same pixels at a different scale. Bicubic upscaling of the whole image narrows the accuracy gap between single-pass inference and multi-turn crop-and-zoom, suggesting that the way pixels are mapped to tokens affects which details the model can use. We trace this dependence at three levels.
\textbf{(1) Within a token.} The vision encoder attenuates high spatial frequencies, making fine patterns with opposite phases harder to distinguish.
\textbf{(2) Across neighboring tokens.} Grid boundaries split patterns, and objects become harder to count separately when they share token positions.
\textbf{(3) Across turns.} Re-encoding a zoomed crop in isolation resets its positional origin and changes its token scale, making it harder to relate the crop to the full scene.
These findings motivate two training-free components. \textbf{Spectrum-Matched Resizing (SMR)} uses the image's spatial spectrum to bring fine detail into the encoder's passband. \textbf{Context-Preserving Inspection (CPI)} returns each enlarged crop together with the full working image in the same turn. \mbox{\textbf{ResVision} (SMR+CPI)} combines the two for agentic inference.
Across 29 MM400 benchmarks ($11{,}221$ items), SMR raises Macro-29 by 0.73--2.66 percentage points over native-resolution Direct inference. On the same metric, ResVision scores 1.50--4.83 points above the existing VISTA baseline, reaching $79.20\%$ on Kimi-K3, $77.84\%$ on Qwen3.8-Flash-Next, $75.45\%$ on GLM-5.3-Flash, and $74.93\%$ on DeepSeek-V4.1-Flash.
\end{abstract}

  \par
  \endgroup

  \vspace{0.65em}
  \noindent
  \begin{minipage}[b]{0.63\linewidth}
    \sffamily\fontsize{8.4}{10.2}\selectfont
    \textbf{Date:} \today\\[-0.1em]
    \textbf{Project Page:} \href{https://pevek.ai}{https://pevek.ai}\\[-0.1em]
    \textbf{Resources:}
    \href{https://github.com/pevek}{GitHub}\,\textperiodcentered\,
    \href{https://huggingface.co/pevek}{HuggingFace}\,\textperiodcentered\,
    \href{https://pevek.ai/blog}{ModelBlog}
  \end{minipage}%
  \hfill
  \begin{minipage}[b]{0.33\linewidth}
    \raggedleft
    \raisebox{-0.30em}{\includegraphics[height=16pt]{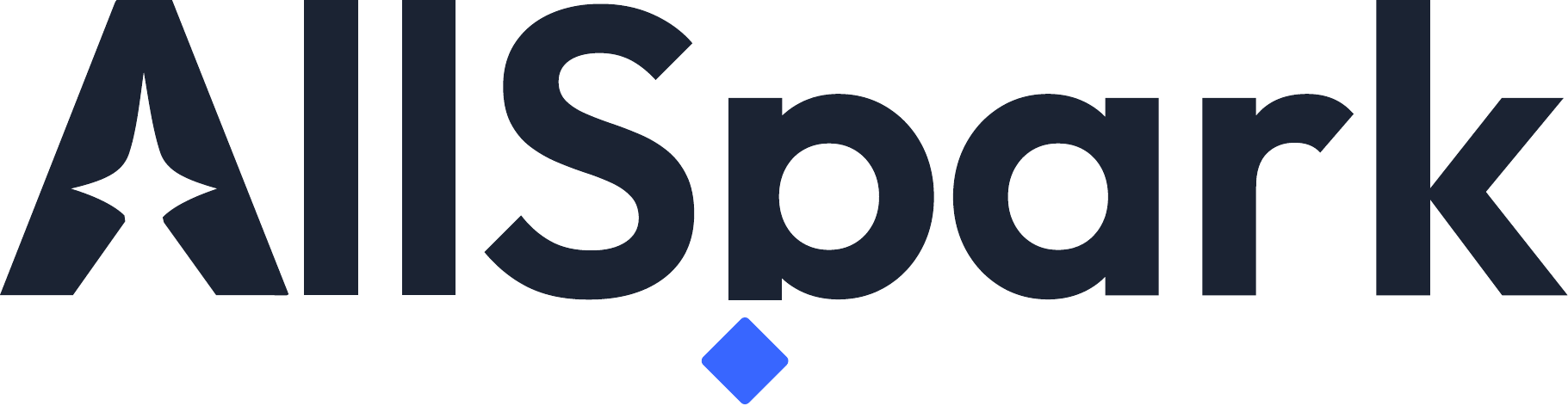}}%
  \end{minipage}
\end{tcolorbox}

\vspace{-0.15em}
\newsavebox{\coverplot}
\sbox{\coverplot}{\resizebox{\linewidth}{!}{
\begingroup
\colorlet{baselinebar}{baselinegray}%
\providecommand{\charttitlefont}{\outfit}%
\providecommand{\chartlabelfont}{\outfit}%
\newcommand{\pairedmodelbars}[8]{%
  \addplot[ybar,bar width=6.0pt,bar shift=-3.4pt,fill=coveraccent,draw=coveraccent,
           nodes near coords={\pgfmathprintnumber[fixed,fixed zerofill,precision=1,assume math mode]{\pgfplotspointmeta}},
           nodes near coords style={font=\chartlabelfont\fontsize{6.6}{7.6}\selectfont,text=abyss,yshift=0.6pt,xshift=-3.15pt}]
           coordinates {#1 #2 #3 #4};
  \addplot[ybar,bar width=6.0pt,bar shift=3.4pt,fill=baselinebar,draw=black!35,
           nodes near coords={\pgfmathprintnumber[fixed,fixed zerofill,precision=1,assume math mode]{\pgfplotspointmeta}},
           nodes near coords style={font=\chartlabelfont\fontsize{6.6}{7.6}\selectfont,text=abyss,yshift=0.6pt,xshift=3.15pt}]
           coordinates {#5 #6 #7 #8};
}%
\newcommand{\comparisonrow}[4]{%
  \node[anchor=base west,inner sep=0pt,font=\charttitlefont\large\bfseries,text=abyss]
    (comparison-heading-#1) at ([yshift=26pt]group c1r#1.north west) {#2};
  \node[anchor=base west,inner xsep=6pt,inner ysep=3pt,
        draw=black!30,fill=white,line width=0.45pt,rounded corners=1.2pt,
        font=\chartlabelfont\normalsize,text=abyss]
    at ([xshift=10pt,yshift=2pt]comparison-heading-#1.base east) {%
      \textcolor{coveraccent}{\rule{9pt}{5pt}}\hspace{4pt}#3\hspace{12pt}%
      {\setlength{\fboxsep}{0pt}\fcolorbox{black!35}{baselinebar}{\rule{0pt}{4.2pt}\rule{8.2pt}{0pt}}}%
      \hspace{4pt}#4};
}%
\begin{tikzpicture}
\begin{groupplot}[
  group style={group size=4 by 2, horizontal sep=0.76cm, vertical sep=3.4cm},
  width=5.86cm, height=4.67cm,
  ybar, axis lines=left, clip=false,
  symbolic x coords={Qwen3.8,Kimi-K3,GLM-5.3,DeepSeek},
  xtick={Qwen3.8,Kimi-K3,GLM-5.3,DeepSeek},
  xticklabels={Qwen3.8,Kimi-K3,GLM-5.3,DeepSeek}, enlarge x limits=0.24,
  xtick align=outside,
  x tick label style={rotate=34, anchor=east, font=\chartlabelfont\footnotesize, text=abyss},
  yticklabel={\pgfmathprintnumber[assume math mode]{\tick}},
  y tick label style={font=\chartlabelfont\footnotesize, text=abyss}, ytick align=outside,
  tick style={black}, axis line style={black},
  title style={font=\charttitlefont\normalsize, yshift=-1pt, text=abyss},
  ymajorgrids, grid style={starlight},
  nodes near coords={\pgfmathprintnumber[fixed,fixed zerofill,precision=1,assume math mode]{\pgfplotspointmeta}},
  every node near coord/.append style={font=\chartlabelfont\scriptsize,text=abyss},
]
\nextgroupplot[
  title={Chart \& OCR}, ymin=80, ymax=88.5
]
  \pairedmodelbars{(Qwen3.8,86.3)}{(Kimi-K3,86.7)}{(GLM-5.3,83.8)}{(DeepSeek,82.8)}{(Qwen3.8,83.7)}{(Kimi-K3,85.5)}{(GLM-5.3,83.8)}{(DeepSeek,82.0)}

\nextgroupplot[
  title={Spatial \& Halluc.}, ymin=66, ymax=78.5
]
  \pairedmodelbars{(Qwen3.8,76.9)}{(Kimi-K3,73.6)}{(GLM-5.3,68.8)}{(DeepSeek,70.8)}{(Qwen3.8,74.1)}{(Kimi-K3,73.4)}{(GLM-5.3,68.7)}{(DeepSeek,68.7)}

\nextgroupplot[
  title={Reason. \& General}, ymin=67, ymax=77.5
]
  \pairedmodelbars{(Qwen3.8,72.0)}{(Kimi-K3,75.5)}{(GLM-5.3,70.0)}{(DeepSeek,71.4)}{(Qwen3.8,69.4)}{(Kimi-K3,74.6)}{(GLM-5.3,68.6)}{(DeepSeek,68.8)}

\nextgroupplot[
  title={Macro-29}, ymin=69, ymax=78.2
]
  \pairedmodelbars{(Qwen3.8,75.5)}{(Kimi-K3,76.5)}{(GLM-5.3,71.6)}{(DeepSeek,72.8)}{(Qwen3.8,72.8)}{(Kimi-K3,75.7)}{(GLM-5.3,70.7)}{(DeepSeek,70.6)}

\nextgroupplot[
  title={Chart \& OCR}, ymin=80, ymax=95.5
]
  \pairedmodelbars{(Qwen3.8,88.4375)}{(Kimi-K3,90.7500)}{(GLM-5.3,86.1250)}{(DeepSeek,85.8125)}{(Qwen3.8,85.3)}{(Kimi-K3,87.8)}{(GLM-5.3,85.5)}{(DeepSeek,82.6)}

\nextgroupplot[
  title={Spatial \& Halluc.}, ymin=66, ymax=82.5
]
  \pairedmodelbars{(Qwen3.8,80.3056)}{(Kimi-K3,76.6111)}{(GLM-5.3,75.1019)}{(DeepSeek,73.7963)}{(Qwen3.8,76.9)}{(Kimi-K3,75.7)}{(GLM-5.3,73.5)}{(DeepSeek,68.9)}

\nextgroupplot[
  title={Reason. \& General}, ymin=65, ymax=80.0
]
  \pairedmodelbars{(Qwen3.8,73.8076)}{(Kimi-K3,77.7773)}{(GLM-5.3,72.9803)}{(DeepSeek,72.8526)}{(Qwen3.8,69.5)}{(Kimi-K3,74.3)}{(GLM-5.3,71.3)}{(DeepSeek,67.7)}

\nextgroupplot[
  title={Macro-29}, ymin=68, ymax=81.0
]
  \pairedmodelbars{(Qwen3.8,77.8421)}{(Kimi-K3,79.2047)}{(GLM-5.3,75.4518)}{(DeepSeek,74.9331)}{(Qwen3.8,74.0)}{(Kimi-K3,76.6)}{(GLM-5.3,74.0)}{(DeepSeek,70.1)}

\end{groupplot}
\comparisonrow{1}{Direct inference}{Direct + SMR}{Native Direct (1×)}
\comparisonrow{2}{Agentic inference}{ResVision}{VISTA}
\end{tikzpicture}
\endgroup
}}
\newlength{\coverplotheight}
\setlength{\coverplotheight}{\dimexpr\pagegoal-\pagetotal-2\intextsep-3\baselineskip-2pt\relax}
\ifdim\coverplotheight>0pt
  \ifdim\coverplotheight<\dimexpr\ht\coverplot+\dp\coverplot\relax
    \sbox{\coverplot}{\resizebox*{!}{\coverplotheight}{\usebox{\coverplot}}}
  \fi
\fi
\begin{figure}[H]
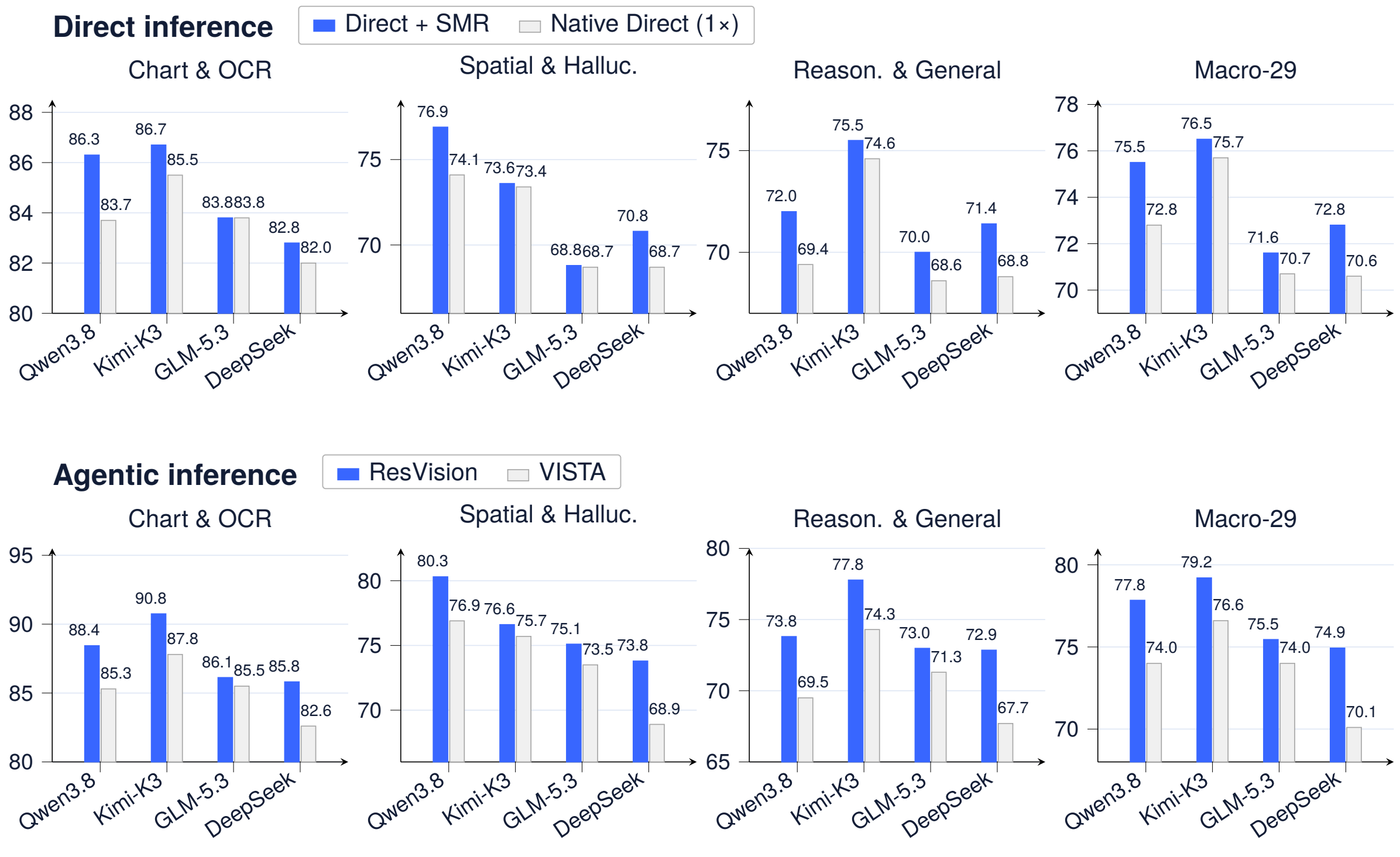

\centering
\usebox{\coverplot}
\captionsetup{font=small}
\caption{\textbf{MM400 accuracy by capability group and overall.} The top row compares Direct + \SMR with native Direct, and the bottom row compares ResVision with VISTA. Scores are unweighted means of benchmark accuracies (\%) within each group (4, 9, and 16 benchmarks) and across all 29 for Macro-29. Axis ranges vary by panel.}
\label{fig:benchmark-comparison}
\end{figure}
\clearpage

\section{Introduction}
\label{sec:intro}

Agentic visual reasoning lets a vision-language model (VLM) choose which parts of an image to inspect before answering a question. A model can survey the scene, zoom in on a chart label or a small object, and use the resulting view to guide its next step~\cite{bajcsy1988active,yao2023react,wu2024vstar,han2026vista}. Each view of a static image resamples evidence that was present from the start. \emph{Why can changing the view make that evidence easier for the model to use?}

\begin{figure*}[htbp]
  \centering
  \input{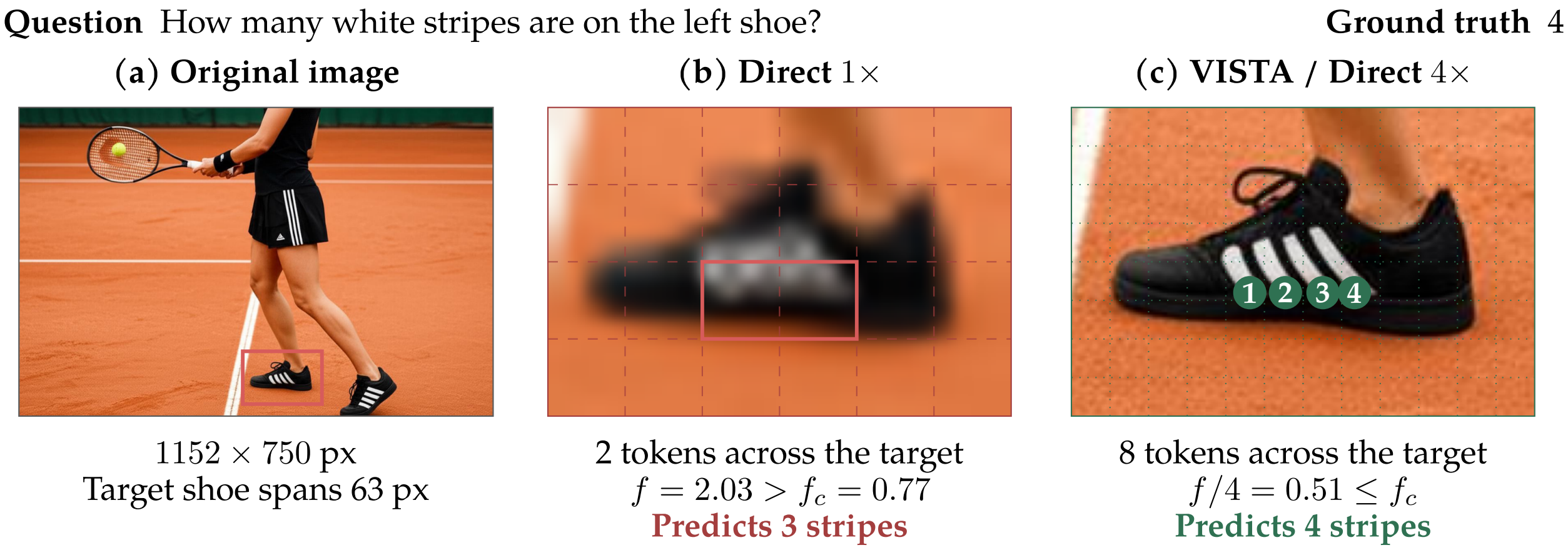}
  \caption{\textbf{An information-complete stripe-counting example from \texttt{VLMsAreBiased}.} Four white stripes are visible on the left shoe in the raw $1152\times 750$~px image. At native scale, \texttt{Direct 1x} represents the logo with $M=2$ visual tokens ($f = 2.03 > f_c = 0.77$~cycles/token, $K=4 > M$) and predicts three stripes. Agentic cropping (\texttt{VISTA}) and global \texttt{Direct 4x} upscaling each expand the logo across $M=8$ tokens ($f/4 = 0.51 \le f_c$) and predict all four stripes.}
  \label{fig:teaser}
\end{figure*}

A simple stripe count shows how much an answer can depend on the view presented to the model. In \cref{fig:teaser}(a), all four white stripes are visible on the left shoe. At native resolution, the stripe pattern spans two visual tokens and the model counts three stripes (\cref{fig:teaser}(b)). In \cref{fig:teaser}(c), an enlarged VISTA crop and a $4\times$ enlargement of the whole photograph each spread the pattern across eight tokens. Both yield the correct count of four. The enlarged full image yields that answer in a single pass. This exposes a gap between \textbf{what the image contains} and \textbf{what the model can recover from its visual tokens}.

A visual token summarizes an image region, so the placement of a detail on the token grid matters. At native resolution, a small cue can share a token with surrounding texture. Enlarging the image can spread that cue across several tokens. With whole-image enlargement, Direct inference applies this finer sampling across the scene in one pass~\cite{liu2023llava,bai2023qwen,chen2024internvl}. Agentic inspection re-encodes just the selected crop, concentrating the extra sampling where the model has chosen to look~\cite{openai2025o3,wang2024qwen2vl}. A smooth photograph and a chart filled with small labels may have identical dimensions and need different sampling densities. \textbf{The working resolution should reflect the image's fine structure.}

When a crop brings a detail into focus, the model still needs to relate it to the full scene. A close-up makes the stripes easier to count, and the full image identifies the left shoe. Surrounding labels, nearby objects, and competing candidates also help the model relate the crop to the question. Re-encoding the crop gives it a new token origin and scale. \textbf{Local inspection should preserve the link between each enlarged detail and its context in the full image.}

We therefore trace evidence at three levels, from what survives within a token to how the model retrieves it across a grid and relates it to the full scene across turns (\cref{sec:analysis}).
\begin{enumerate}
  \item \textbf{Single-Token Bandwidth Limit (\cref{sec:intra_token}):} The vision encoder attenuates high-frequency detail before the image tokens reach the language model. The measured cutoff is $f_c = 0.77$~cycles/token.
  \item \textbf{Contiguous-Token Retrieval Breakdown (\cref{sec:inter_token}):} Grid boundaries can split patterns and weaken their features. When objects outnumber tokens ($K > M$), our probe of two-dimensional rotary position embeddings (2D RoPE) shows peak attention settling on one column before the count is complete.
  \item \textbf{Separated-Token Context Detachment (\cref{sec:disjoint_tokens}):} Re-encoding an isolated crop resets its 2D RoPE origin and changes its scale. The model must then match the local evidence back to the full scene.
\end{enumerate}

\begin{samepage}
These findings guide two decisions in \textbf{ResVision (SMR+CPI)} (\cref{sec:method}). The first is how to sample the full image. \textbf{Spectrum-Matched Resizing (SMR)} estimates the image spectrum and uses the measured encoder cutoff to set a working resolution within a visual-token budget. The model can answer from this working view or select a region for closer inspection. The second decision is what the model receives when it requests a crop. \textbf{Context-Preserving Inspection (CPI)} pairs the enlarged crop with the complete working image and marks where the crop came from. Together, the two training-free components improve the detail available at the start and preserve a full-image reference as the model inspects individual regions.

\end{samepage}

We evaluate the approach on MM400, which spans 29 benchmarks and $11{,}221$ items, and use a focused five-benchmark suite to study resizing, inspection, and token allocation under a common tool-call limit for agentic methods. On MM400, SMR improves Direct Macro-29 by 0.73--2.66 points across \texttt{Qwen3.8-Flash-Next}~\cite{qwen2026design}, \texttt{GLM-5.3-Flash}~\cite{zai2026flash}, \texttt{DeepSeek-V4.1-Flash}~\cite{deepseek2026flash}, and \texttt{Kimi-K3}~\cite{moonshot2026kimi}. For the same four models, ResVision scores 1.50--4.83 Macro-29 points above the existing VISTA baseline.

\section{Understanding the Bandwidth of Image Tokens in VLMs}
\label{sec:analysis}

\begin{figure*}[!htbp]
  \centering
  \begingroup
\tikzset{
  bandwidth/panel/.style={rounded corners=3pt,draw=black!20,fill=ablmodelband!38,line width=.55pt},
  bandwidth/title/.style={font=\fontsize{10.2}{12}\selectfont\bfseries,anchor=west,text=black},
  bandwidth/label/.style={font=\fontsize{9}{11}\selectfont,align=center,text=black},
  bandwidth/arrow/.style={-{Latex[length=1.55mm]},draw=black!85,line width=.68pt},
  bandwidth/grid/.style={draw=ablglobaledge,line width=.65pt},
  bandwidth/limit/.style={draw=abldirectedge,line width=.9pt}}
\resizebox{\textwidth}{!}{%
\begin{tikzpicture}[x=1cm,y=1cm]
  \path[use as bounding box] (0,0) rectangle (16.45,7.36);

  \begin{scope}[yshift=4.96cm]
    \draw[bandwidth/panel] (.03,.03) rectangle (16.42,2.37);
    \node[bandwidth/title] at (.25,1.49) {(a) Single token};
    \node[bandwidth/label,anchor=west,align=left] at (.25,.90)
      {Low-pass filtering\\Weakens fine detail};
    \draw[bandwidth/grid,fill=white] (4.14,.88) rectangle (5.35,2.09);
    \foreach \x in {4.25,4.42,4.59,4.76,4.93,5.10,5.27}
      \draw[black,line width=.65pt] (\x,.96)--(\x,2.01);
    \draw[ablglobaledge,dashed] (4.745,.88)--(4.745,2.09) (4.14,1.485)--(5.35,1.485);
    \node[bandwidth/label] at (4.745,.48) {$32\times32$ px\\$2\times2$ patches};
    \draw[bandwidth/arrow] (5.52,1.48)--(6.45,1.48);
    \draw[bandwidth/grid,fill=white,rounded corners=2pt] (6.61,.38) rectangle (10.11,2.12);
    \node[bandwidth/label] at (8.36,1.87) {ViT + $2\times2$ merger};
    \fill[ablglobalband] (7.05,.88) rectangle (8.16,1.58);
    \draw[bandwidth/arrow] (7.05,.88)--(9.83,.88);
    \draw[bandwidth/arrow] (7.05,.88)--(7.05,1.60);
    \draw[abldirectedge,dashed] (8.16,.88)--(8.16,1.58);
    \draw[ablglobaledge,line width=.9pt] (7.05,1.51) .. controls (7.62,1.51) and (7.93,1.42) .. (8.16,1.22)
      .. controls (8.53,.95) and (9.03,.94) .. (9.70,.94);
    \node[bandwidth/label] at (8.36,.62) {$f_c=0.77$ cycles/token};
    \draw[bandwidth/arrow] (10.27,1.48)--(10.99,1.48);
    \draw[bandwidth/grid,fill=black!23] (11.16,.99) rectangle (12.14,1.97);
    \node[bandwidth/label] at (11.65,1.48) {$\mathbf{z}_{r,c}$};
    \node[bandwidth/label,anchor=west,align=left] at (12.40,1.48)
      {Phase reversal\\Cosine distance $0.069$\\at $f=2$ cycles/token};
  \end{scope}

  \begin{scope}[yshift=2.48cm]
    \draw[bandwidth/panel] (.03,.03) rectangle (16.42,2.37);
    \node[bandwidth/title] at (.25,1.49) {(b) Contiguous tokens};
    \node[bandwidth/label,anchor=west,align=left] at (.25,.90)
      {Grid splitting and\\counting saturation};
    \draw[bandwidth/grid,fill=white] (4.20,.94) rectangle (7.62,1.77);
    \draw[fill=ablglobalband,draw=ablglobaledge] (4.87,1.355) circle (.19);
    \draw[fill=abldirect!27,draw=abldirectedge] (5.91,1.355) circle (.25);
    \draw[fill=ablglobalband,draw=ablglobaledge] (6.95,1.355) circle (.19);
    \draw[bandwidth/limit] (5.91,.86)--(5.91,1.85);
    \node[bandwidth/label] at (5.91,.49)
      {Split at $x=32$ px\\Feature contrast falls by 57\%};
    \draw[bandwidth/grid,fill=white] (9.31,.94) rectangle (13.99,1.77);
    \foreach \x in {10.48,11.65,12.82}
      \draw[bandwidth/grid] (\x,.94)--(\x,1.77);
    \foreach \x/\stepno in {9.895/1,11.065/2,12.235/3,13.405/4} {
      \fill[ablglobaledge] (\x,1.22) circle (.115);
      \node[bandwidth/label] at (\x,2.00) {$k=\stepno$};
    }
    \foreach \x in {10.77,13.72}
      \fill[abldirectedge] (\x,1.22) circle (.115);
    \draw[bandwidth/arrow] (9.52,1.53)--(13.83,1.53);
    \draw[bandwidth/limit] (13.99,.86)--(13.99,1.85);
    \node[bandwidth/label] at (12.28,.49)
      {2D RoPE pointer stops at the grid edge\\At $K>M$: ViT probe 100\%, LLM counting 0\%};
  \end{scope}

  \begin{scope}
    \draw[bandwidth/panel] (.03,.03) rectangle (16.42,2.37);
    \node[bandwidth/title] at (.25,1.49) {(c) Separated tokens};
    \node[bandwidth/label,anchor=west,align=left] at (.25,.90)
      {RoPE reset and\\scale mismatch};
    \fill[white] (4.08,.77) rectangle (5.84,2.09);
    \draw[step=.44,draw=ablglobaledge!35,line width=.42pt,shift={(4.08,.77)}] (0,0) grid (1.76,1.32);
    \draw[bandwidth/grid] (4.08,.77) rectangle (5.84,2.09);
    \draw[black!55,line width=.6pt] (4.65,1.91)--(5.62,1.91) (4.26,.90)--(4.73,.90);
    \draw[bandwidth/limit,fill=abldirect!13] (4.96,1.21) rectangle (5.40,1.65);
    \draw[ablglobaledge] (5.05,1.29)--(5.18,1.55)--(5.31,1.29)--cycle;
    \node[bandwidth/label] at (4.96,.44) {Turn 0: full image, stride $p$};
    \draw[bandwidth/arrow] (6.03,1.36)--node[bandwidth/label,above=2pt] {Crop and zoom $s\times$} (8.85,1.36);
    \node[bandwidth/label] at (7.44,1.04) {Across tool turns};
    \fill[white] (9.02,.77) rectangle (10.78,2.09);
    \draw[step=.44,draw=ablglobaledge!35,line width=.42pt,shift={(9.02,.77)}] (0,0) grid (1.76,1.32);
    \draw[bandwidth/limit] (9.02,.77) rectangle (10.78,2.09);
    \draw[ablglobaledge,line width=.75pt] (9.53,.95)--(9.95,1.79)--(10.37,.95)--cycle;
    \node[bandwidth/label,anchor=north west,fill=white,inner sep=1pt] at (9.05,2.05) {$(0,0)$};
    \node[bandwidth/label] at (9.90,.44) {Turn $t$: crop, stride $p/s$};
    \node[bandwidth/label,anchor=west,align=left] at (11.23,1.43)
      {Target starts at $(r_0,c_0)$ in Turn 0\\Crop resets the origin to $(0,0)$\\Surrounding landmarks are lost};
  \end{scope}
\end{tikzpicture}%
}
\endgroup
  \caption{\textbf{Three image-token bottlenecks in VLMs.} \textbf{(a)} Fine spatial phase is attenuated within one token (\cref{sec:intra_token}). \textbf{(b)} Grid boundaries split patterns. Counting degrades when objects outnumber tokens ($K>M$, \cref{sec:inter_token}). \textbf{(c)} Isolated crops disrupt spatial and visual matching across turns (\cref{sec:disjoint_tokens}).}
  \label{fig:sec2_overview}
\end{figure*}
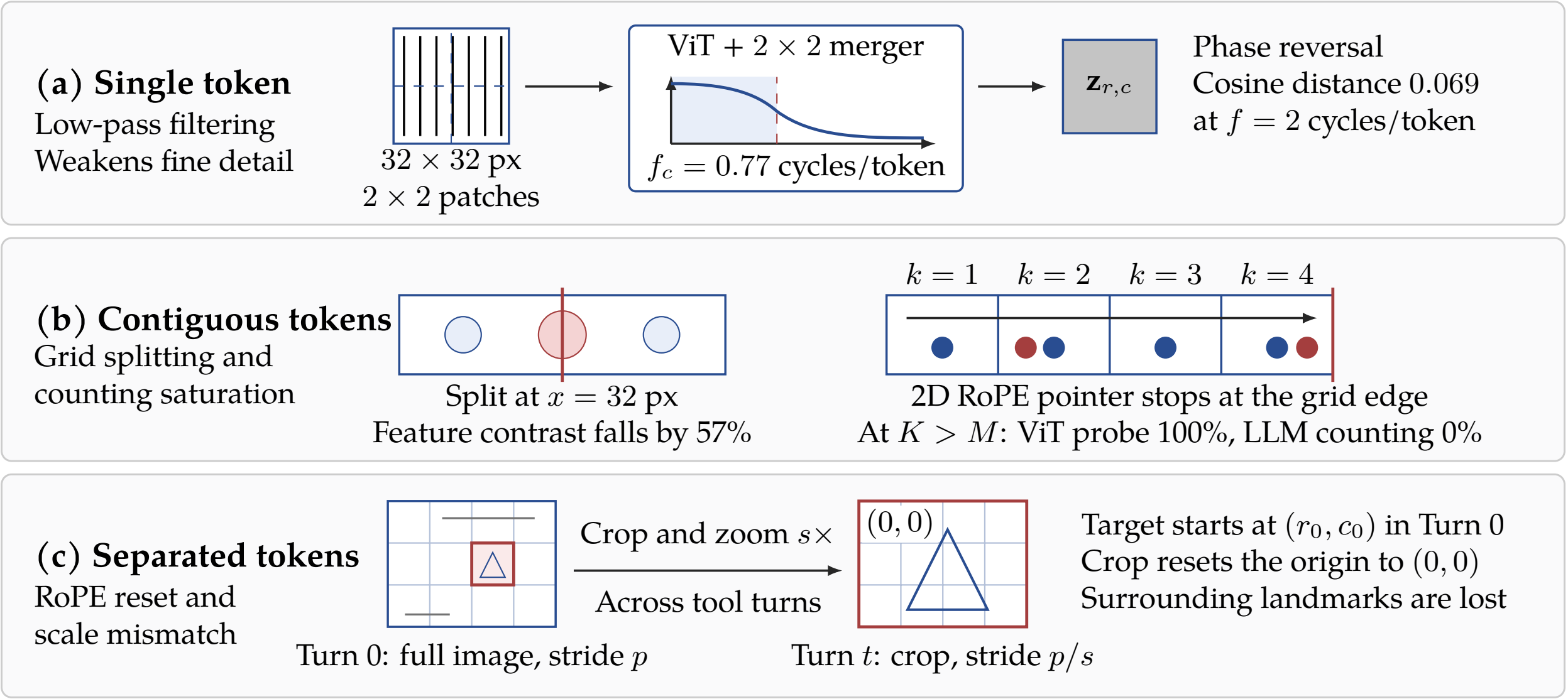

To understand why the same pixels can yield different answers, we follow visual evidence from the encoder to the inspection loop. A detail must survive encoding, remain distinguishable among neighboring tokens, and stay linked to the full scene when the model inspects it in a later turn (\cref{fig:sec2_overview}).

The encoder we probe is Qwen3.8. It converts an image into visual tokens in three stages.
\begin{itemize}
  \item \textbf{Stage 1: Patch Projection.} A strided convolution divides the image into non-overlapping $16\times16$~px patches and projects each patch into an embedding.
  \item \textbf{Stage 2: Vision Transformer (ViT) Blocks.} A stack of self-attention blocks with 2D RoPE processes the patch tokens at full patch resolution.
  \item \textbf{Stage 3: Spatial Patch Merger.} Each $2\times 2$ neighborhood of Stage-2 tokens (covering $32\times 32$~px) is concatenated and projected by a two-layer MLP into a single visual token $\mathbf{z}_{r,c}$ with a discrete integer 2D RoPE coordinate $(r, c)$.
\end{itemize}

\subsection{Single Image Token: Low-Pass Filtering}
\label{sec:intra_token}

\begin{figure*}[htbp]
  \centering
  \begin{subfigure}[b]{0.325\textwidth}
    \centering
    \includegraphics[width=\linewidth]{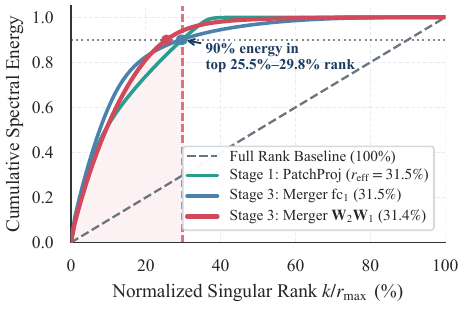}
    \caption{Filter SVD cumulative energy.}
    \label{fig:intra_a}
  \end{subfigure}
  \hfill
  \begin{subfigure}[b]{0.325\textwidth}
    \centering
    \includegraphics[width=\linewidth]{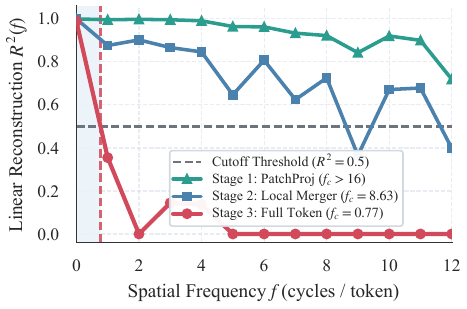}
    \caption{3-stage linear reconstruction $R^2(f)$.}
    \label{fig:intra_b}
  \end{subfigure}
  \hfill
  \begin{subfigure}[b]{0.325\textwidth}
    \centering
    \includegraphics[width=\linewidth]{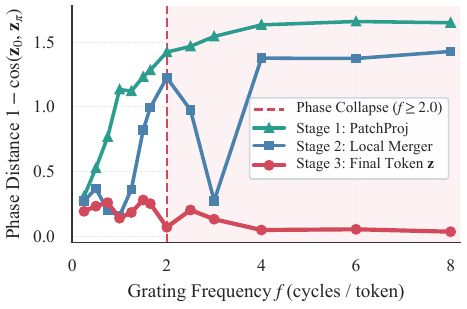}
    \caption{Phase-reversal test ($\phi=0$ vs.\ $\pi$).}
    \label{fig:intra_c}
  \end{subfigure}
  \caption{\textbf{Single-token low-pass filtering characteristics (\cref{sec:intra_token}).} (\subref{fig:intra_a}) Cumulative singular value energy of Stage-1 and Stage-3 weight matrices, where effective rank contracts to ${\sim}31.5\%$ and the top $25.5\%$--$29.8\%$ of singular vectors account for $90\%$ of the energy. (\subref{fig:intra_b}) Linear reconstruction $R^2(f)$ across Stages 1--3, showing that the $R^2=0.50$ cutoff drops from the $16$~cycles/token sampling limit (Stage 1) and $8.63$~cycles/token (isolated local merger) to $f_c = 0.77$~cycles/token after the full encoder (Stage 3). (\subref{fig:intra_c}) Phase-reversal test ($\phi=0$ vs.\ $\phi=\pi$), where final tokens are less sensitive to opposite phases than patch embeddings.}
  \label{fig:intra_mtf}
\end{figure*}

\paragraph{Theoretical Capacity vs.\ Filter Spectrum.}
A $32\times32$~px region contains four $16\times16$ RGB patches, or $3{,}072$ scalar inputs, with a Nyquist limit of $16$~cycles per $32$~px along each axis. The final visual token has $2{,}048$ dimensions. To characterize the learned weights, we examine the singular value spectra of the patch projection and merger in \cref{fig:intra_a}. The patch projection has a Shannon effective rank of $241.8$ out of $768$, and the first merger layer has an effective rank of $1{,}449.7$ out of $4{,}608$, both about $31.5\%$ of their maximum matrix rank. This concentration motivates measuring frequency response directly.

\paragraph{Frequency Cutoff Across Encoding Stages.}
A frequency of two cycles per token means that two repetitions of a pattern fit across one token's spatial footprint. We fit linear ridge decoders to reconstruct the $32\times32$~px input at different spatial frequencies, using $2{,}560$ regions for training and $512$ for testing (\cref{fig:intra_b}). We define $f_c$ by the first crossing of $R^2=0.50$. Patch projection retains high reconstruction accuracy up to the $16$~cycles/token sampling limit, and the isolated local merger control gives $f_c=8.63$~cycles/token. With the ViT blocks and merger, \textbf{the full encoder has a cutoff of $0.77$~cycles/token}, corresponding to a wavelength of about $41.6$~px.

\paragraph{Phase-Reversal Test.}
We test sensitivity to high-frequency phase by presenting $32\times32$ sine-wave gratings that differ by a phase shift of $\pi$. We measure the cosine distance $1-\cos(\mathbf{z}_{0},\mathbf{z}_{\pi})$ between the resulting features (\cref{fig:intra_c}). At $2$~cycles/token, this distance is $1.4231$ after patch projection and $0.0687$ in the final visual tokens. A fine pattern can therefore change while producing nearly the same final token.

\subsection{Contiguous Image Tokens: Grid Splitting and Counting Limit}
\label{sec:inter_token}

\begin{figure*}[htbp]
  \centering
  \begin{subfigure}[b]{0.325\textwidth}
    \centering
    \includegraphics[width=\linewidth]{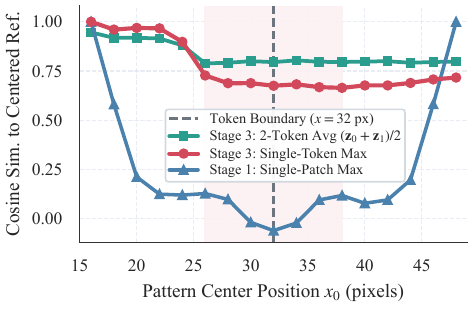}
    \caption{Cross-boundary splitting ($x=32$~px).}
    \label{fig:inter_a}
  \end{subfigure}
  \hfill
  \begin{subfigure}[b]{0.325\textwidth}
    \centering
    \includegraphics[width=\linewidth]{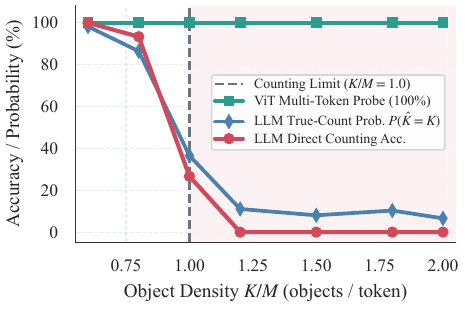}
    \caption{ViT probe vs.\ LLM counting ($M=10$).}
    \label{fig:inter_b}
  \end{subfigure}
  \hfill
  \begin{subfigure}[b]{0.325\textwidth}
    \centering
    \includegraphics[width=\linewidth]{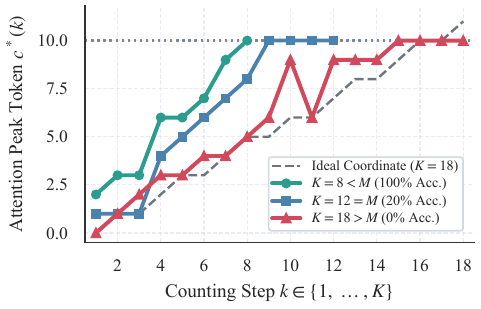}
    \caption{2D RoPE pointer saturation ($M=12$).}
    \label{fig:inter_c}
  \end{subfigure}
  \caption{\textbf{Contiguous image token bottlenecks (\cref{sec:inter_token}).} (\subref{fig:inter_a}) Sliding a $14\times 14$~px pattern across the $x=32$~px boundary between two adjacent tokens reduces Stage-3 single-token similarity by $32.5\%$ and odd-one-out feature contrast by $57.0\%$. (\subref{fig:inter_b}) Across $M=10$ contiguous tokens, a linear probe on Stage-3 ViT tokens achieves $100.0\%$ counting accuracy for the tested object counts from $6$ to $20$, whereas LLM counting collapses from $93.3\%$--$100.0\%$ ($K < M$) to $0.0\%$ for the tested counts with $K > M$. (\subref{fig:inter_c}) Peak attention column across $M=12$ tokens during step-by-step counting. With $K=18$ objects, attention saturates at column $c^*=10$.}
  \label{fig:inter_phase}
\end{figure*}

The single-token probes measure whether a pattern survives encoding. Its placement on the grid adds a second constraint. We move patterns across token boundaries, then ask how well the model counts when several objects share the same tokens (\cref{fig:inter_phase}).

\paragraph{Pattern Splitting at Token Boundaries.}
Because the $32\times 32$~px token grid cuts the image at fixed $32$~px intervals, repeated objects often fall right on a boundary between two tokens. As shown in \cref{fig:inter_a}, sliding a $14\times 14$~px pattern from Token $0$'s center ($x = 16$~px) across the $x=32$~px boundary to Token $1$'s center ($x = 48$~px) causes Stage-1 single-token feature similarity to drop from $1.00$ to $-0.06$, and Stage-3 similarity to drop from $1.00$ to $0.68$ ($-32.5\%$), whereas averaging both adjacent tokens recovers $0.80$ similarity. When three objects sit inside two tokens ($x \in \{11, 32, 53\}$~px), the middle object is cut in half by the $x=32$~px boundary, reducing its feature distinction from neighboring objects by \textbf{$57.0\%$}. The model must integrate evidence from both token positions to distinguish the middle object.

\paragraph{Counting When Objects Share Tokens.}
When multiple objects appear in a row, does counting fail because the ViT encoder blurs them together, or because the LLM cannot index them? We test this by placing $K \in \{6, 8, 10, 12, 15, 18, 20\}$ clear shapes ($10$--$12$~px each) inside a fixed strip of $M = 10$ horizontal tokens ($320\times 64$~px, \cref{fig:inter_b}). A linear classifier reads the count from the 10 Stage-3 ViT tokens with $100.0\%$ accuracy. The LLM achieves $93.3\%$--$100.0\%$ for $K<M$, drops to $26.7\%$ at $K=M$, and scores $0.0\%$ for the tested counts with $K>M$. \textbf{The count is still recoverable from the visual tokens when the LLM fails to report it.}

Tracking which visual token column receives peak attention at each counting step $k = 1, \dots, K$ across $M=12$ tokens (\cref{fig:inter_c}) helps explain this result. Since 2D RoPE gives one integer coordinate $(r, c)$ to each $32\times 32$ token, the LLM's attention shifts by roughly one token column per counted number. When $K = 18$ objects are packed into $M = 12$ tokens, peak attention reaches column $c^* = 10$ at step $k = 15$ and remains there through the end of the trace. The plateau suggests that the model's token-by-token counting strategy breaks down when several objects share a position on the token grid.

\subsection{Separated Image Tokens: 2D RoPE Reset and Zoom Scale Mismatch}
\label{sec:disjoint_tokens}

\begin{figure*}[htbp]
  \centering
  \begin{subfigure}[b]{0.325\textwidth}
    \centering
    \includegraphics[width=\linewidth]{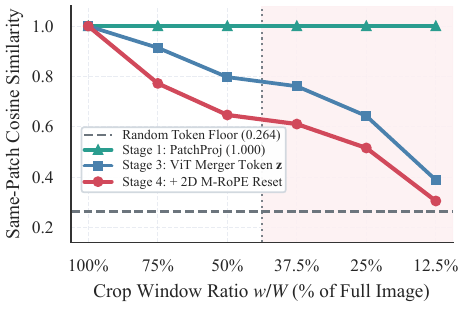}
    \caption{Same-patch $\cos$ across crop ratios.}
    \label{fig:sep_a}
  \end{subfigure}
  \hfill
  \begin{subfigure}[b]{0.325\textwidth}
    \centering
    \includegraphics[width=\linewidth]{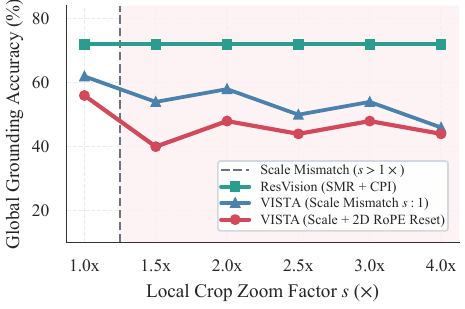}
    \caption{Grounding vs.\ zoom scale $s$.}
    \label{fig:sep_b}
  \end{subfigure}
  \hfill
  \begin{subfigure}[b]{0.325\textwidth}
    \centering
    \includegraphics[width=\linewidth]{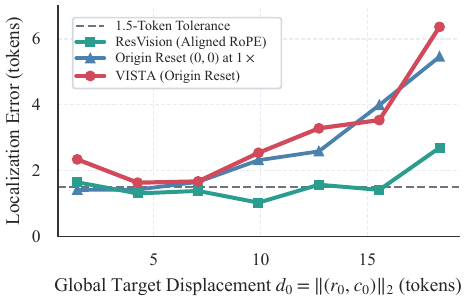}
    \caption{Localization error vs.\ offset $d_0$.}
    \label{fig:sep_c}
  \end{subfigure}
  \caption{\textbf{Separated image token bottlenecks in multi-turn agentic cropping (\cref{sec:disjoint_tokens}).} (\subref{fig:sep_a}) Cropping around the exact same $32\times 32$~px target patch at $s=1\times$ preserves Stage-1 similarity ($\cos = 1.000$), but Stage-3 token similarity drops to $0.389$ ($0.305$ with 2D M-RoPE reset) as the crop side length shrinks to $12.5\%$ of the original, reducing top-1 patch retrieval from $100.0\%$ to $28.0\%$. (\subref{fig:sep_b}) Across tested zoom factors from $1.5\times$ to $4\times$, 2D RoPE scale mismatch after zooming and origin reset reduce \texttt{VISTA} global grounding accuracy to $40.0\%$--$48.0\%$, whereas \mbox{\texttt{ResVision} (SMR+CPI)} maintains $72.0\%$. (\subref{fig:sep_c}) For the $2\times$ isolated crop with origin reset, localization error rises from $2.34$ to $6.37$~tokens as target displacement increases from $1.41$ to $18.38$~tokens.}
  \label{fig:sep_probe}
\end{figure*}

Enlarging a region spreads fine patterns and crowded objects over more tokens, addressing the two constraints above. Agentic methods such as \texttt{VISTA}~\cite{han2026vista} do this by re-encoding a crop in a later turn. The model then has to match the enlarged detail to its location in the original scene. We separate the effects of re-encoding the crop in isolation and enlarging it before that match (\cref{fig:sep_probe}).

\begin{itemize}
  \item \textbf{2D RoPE Origin Reset and Context Truncation (\cref{fig:sep_a,fig:sep_c}):} Within a single contiguous image (Turn $0$), a sub-region located at token offset $(r_0, c_0)$ carries its global 2D RoPE coordinates $(r_0 + \Delta r, c_0 + \Delta c)$. When that sub-region is cropped and re-encoded as a standalone image in Turn $t$, the vision encoder truncates self-attention to the crop boundary and resets its 2D RoPE coordinates to start from $(0, 0)$ at the crop's top-left corner:
  \begin{equation}
    (r', c')_{\mathrm{crop}} \in [0, H'/p) \times [0, W'/p) \;\neq\; (r_0 + \Delta r,\; c_0 + \Delta c)_{\mathrm{global}}.
    \label{eq:rope_reset}
  \end{equation}
  To quantify this effect across $50$ BabyVision images ($512\times 512$~px, $16\times 16$ tokens), we extract $32$~px-aligned crops around the \emph{exact same} target patch at $s=1\times$ (\cref{fig:sep_a}). While Stage-1 local convolution preserves feature direction ($\cos = 1.000$), shrinking the crop side length from $100\%$ to $12.5\%$ of the original ($2\times 2$ tokens) causes Stage-3 token cosine similarity to drop to \textbf{$0.389$} (and to \textbf{$0.305$} under LLM 2D M-RoPE reset---within $0.041$ of the \textbf{$0.264$} cosine similarity floor between random unrelated foreground tokens), reducing top-1 patch retrieval accuracy from $100.0\%$ to \textbf{$28.0\%$}. Localization errors for isolated crops grow as targets move farther from the image origin (\cref{fig:sep_c}).
  \item \textbf{2D RoPE Scale Mismatch After Zooming (\cref{fig:sep_b}):} When the cropped region is enlarged by a factor $s > 1$, the same visual object now spans $s\times$ more tokens along each axis than it did in Turn $0$. As a result, its internal 2D RoPE coordinate distances are stretched by $s\times$ while the unzoomed Turn-0 image remains blurred at $1\times$ resolution, making it difficult for the model to recognize that the zoomed crop and the original Turn-0 region correspond to the same pattern. Across zoom factors $s \in [1.5\times, 4.0\times]$ (\cref{fig:sep_b}), matching the zoomed crop back to the $1\times$ Turn-0 image reduces cross-turn token similarity from $0.600$ to \textbf{$0.364$--$0.399$} and drops global grounding accuracy to \textbf{$40.0\%$--$48.0\%$}, compared to \textbf{$72.0\%$} when keeping a unified scale and in-turn context via \texttt{ResVision}.
\end{itemize}

Together, the probes point to two design choices. The working resolution should make fine detail readable on the token grid, and the inspection reply should carry the global reference that locates the detail in the scene. \Cref{sec:method} develops both choices.

\section{Methodology}
\label{sec:method}

ResVision makes two decisions about how the model sees an image. \textbf{\SMR} sets the working resolution from the image spectrum so the first view captures more of the fine structure. If the model asks for a closer look, \textbf{CPI} returns the enlarged crop alongside the full working image and marks where it came from. Both components operate at inference time and share the same working image (\cref{sec:method_agentic}).

\subsection{Direct Response via \SMR}
\label{sec:method_direct}

We compare two ways to choose a working resolution. A fixed budget gives a smooth photograph and a crowded diagram the same number of tokens. \SMR can sample the diagram more finely when the spacing of its marks matters.

\paragraph{Policy 1: Fixed Token Budget (\texttt{fixed\_tokens}).}
At native resolution, an $800\times800$~px diagram occupies about 625 tokens at a stride of $32$~px. Dense diagrams can place several objects within the footprint of one token. A fixed target increases sampling for such images by resizing each input to the same token count. Given input dimensions $(W,H)$ and a target $N_{\mathrm{target}}=4096$, we scale isotropically by
\begin{equation}
  s_{\mathrm{fixed}} = \sqrt{\frac{1024 \cdot N_{\mathrm{target}}}{W \cdot H}},
  \label{eq:fixed_tokens}
\end{equation}
rounding dimensions to multiples of $32$~px, using bicubic interpolation when $s_{\mathrm{fixed}} \ge 1$ and Lanczos filtering when $s_{\mathrm{fixed}} < 1$.

\paragraph{Policy 2: \SMR.}
Enlarging an image spreads a pattern across more tokens and lowers its frequency in cycles per token. \SMR estimates how far the image's spectral energy extends and uses the measured cutoff $f_c = 0.77$~cycles/token (\cref{sec:intra_token}) to set a target scale.
\begin{enumerate}
  \item \textbf{Estimate local spectral power.} Convert the image to grayscale and keep the highest-variance quarter of its $64\times64$~px blocks, so that blank areas contribute less to the scale estimate. Subtract the mean of each selected block to obtain $I_{0,b}$ and apply a 2D Hann window $w_{\mathrm{Hann}}$ to reduce boundary leakage. The block's spectral power is
  \[
    S_b(f_y,f_x) = \left|\mathcal{F}_{2\mathrm{D}}\!\left(I_{0,b}\odot w_{\mathrm{Hann}}\right)(f_y,f_x)\right|^2,
    \qquad f_x,f_y\in[-0.5,0.5]\ \text{cycles/px}.
  \]

  \item \textbf{Find a characteristic frequency.} Average the block spectra and pool their power into rings at radial frequency $f=\sqrt{f_x^2+f_y^2}$ over $(0,0.5]$~cycles/px. The radial spectrum $P(f)$ pools energy across orientations. We interpolate its cumulative energy curve to find the quantile $f_\eta$:
  \[
    \sum_{f\leq f_\eta} P(f) = \eta \sum_f P(f).
  \]
  We use $\eta=0.90$ by default (\cref{sec:suppl_ablation}).

  \item \textbf{Map frequency to a resize scale.} With token stride $p=32$~px, the frequency $f_\eta$ maps to $p f_\eta/s$ cycles per token at scale $s$. The smallest scale that brings it to the cutoff is $p f_\eta/f_c$. We clip this scale to the limits implied by the token budget $[N_{\min},N_{\max}]$.
  \begin{equation}
    \begin{gathered}
      s_{\mathrm{spec}} = \operatorname{clip}\!\left(\frac{p f_{\eta}}{f_c},\,s_{\min},\,s_{\max}\right),\\
      s_{\min} = \sqrt{\frac{N_{\min}p^2}{WH}},
      \qquad
      s_{\max} = \sqrt{\frac{N_{\max}p^2}{WH}}.
    \end{gathered}
    \label{eq:s_spec}
  \end{equation}
\end{enumerate}

The cutoff sets the desired scale, and the token budget bounds the full-image resize. A later crop can devote more tokens to a region that still needs inspection.

\subsection{Agentic Perception via ResVision}
\label{sec:method_agentic}

\begin{figure*}[htbp]
  \centering
  \input{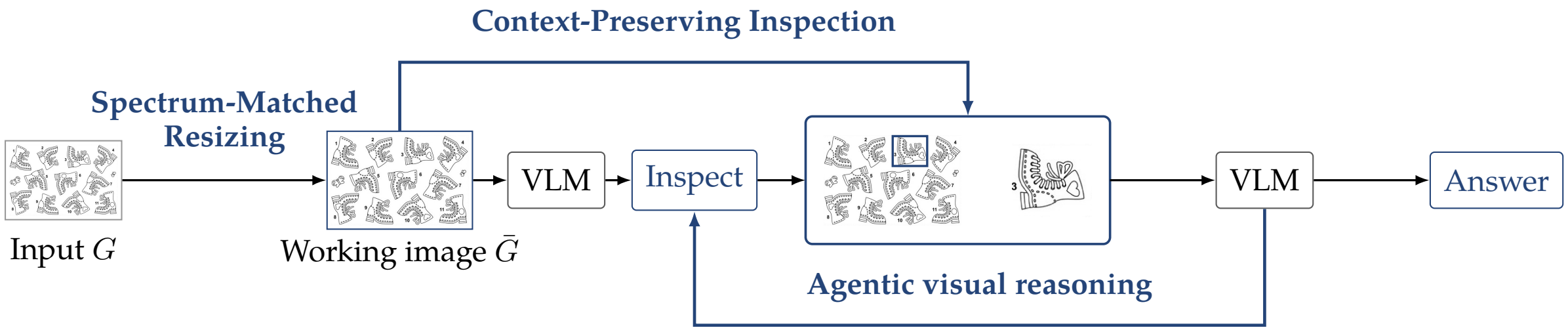}
  \caption{\textbf{Overview of \mbox{ResVision (SMR+CPI)} for multi-turn agentic visual reasoning.} \SMR sets the working image $\bar{G}$ before the first model call. Each CPI reply returns the enlarged crop $Z_{L_t}$ and the working view $Z_{\bar{G}}$ with the crop location marked.}
  \label{fig:resvision_overview}
\end{figure*}

A crop can make a mark legible while obscuring which object, row, or candidate it belongs to (\cref{sec:disjoint_tokens}). ResVision gives the model a common reference for the two views. In \cref{fig:resvision_overview}, the enlarged detail shows the pattern, and the box on the working image identifies the shoe it belongs to. This correspondence depends on both visual features and token positions. Zoom changes the crop's token spacing to $p/s$ in source-image coordinates, and intervening text adds a positional offset in the conversation.

\paragraph{1. Reducing Global--Local Feature Discrepancy via \SMR.}
Before the first model call, ResVision applies \cref{eq:s_spec} to form the working image $\bar{G}=U_{s_{\mathrm{spec}}}(G)$. We align its dimensions to the $32$~px token grid and share the visual-token allowance across images in multi-image requests. The working image serves as both the initial global view and the source of later crops. Starting with a more detailed global view makes its features easier to match with those of the crop.

\paragraph{2. Reducing Cross-Turn Positional Separation via CPI.}
When the VLM selects a bounding box $B_t$ at step $t$ to inspect an ambiguous region, the tool extracts and enlarges the local view from the working image:
\begin{equation}
  L_t = U_{\mathrm{local}} C_{B_t}(\bar{G}),
  \label{eq:local_view}
\end{equation}
where $U_{\mathrm{local}}$ is the local detail renderer.

An isolated crop has its own image grid and appears later in the conversation. In 3D M-RoPE, the intervening text advances the crop's starting index by $\Delta t_{\mathrm{seq}}$ in each of the temporal, height, and width dimensions. The position of a crop token is
\[
  (T,H,W)_{\mathrm{crop}} =
  \bigl(t_0+\Delta t_{\mathrm{seq}},\;
        t_0+\Delta t_{\mathrm{seq}}+r_{\mathrm{loc}},\;
        t_0+\Delta t_{\mathrm{seq}}+c_{\mathrm{loc}}\bigr),
\]
where $r_{\mathrm{loc}}$ and $c_{\mathrm{loc}}$ are the token's row and column within the crop.

\begin{wrapfigure}{R}{0.48\textwidth}
  \vspace{-8pt}
  \centering
  \includegraphics[width=\linewidth]{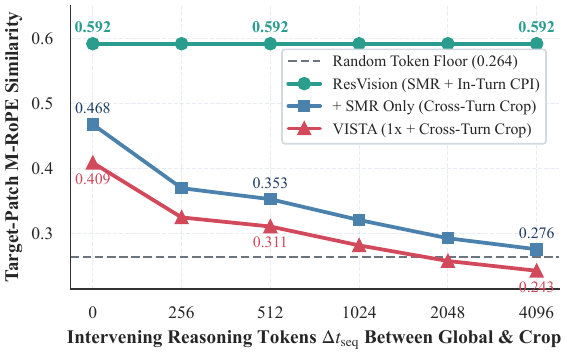}
  \caption{\textbf{Target-patch 3D M-RoPE similarity across intervening reasoning length $\Delta t_{\mathrm{seq}}$ ($N=50$ BabyVision images).} 
  }
  \label{fig:mrope_cross_turn}
  \vspace{-8pt}
\end{wrapfigure}
An isolated crop moves farther from the original view in M-RoPE coordinates as reasoning continues. With 4,096 intervening text tokens, target-patch similarity drops from $0.409$ to $0.243$ for \texttt{VISTA}, below the $0.264$ random-token floor, and from $0.468$ to $0.276$ with \SMR alone (\cref{fig:mrope_cross_turn}). CPI brings the full working image into the current reply, so the model can match the local detail to a view in the same turn. We write the reply as $M_t$ and the accumulated visual evidence as $E_t$:
\begin{equation}
  \begin{aligned}
    M_t &= Z_{\bar{G}} \oplus Z_{L_t}, \\
    E_0 &= Z_{\bar{G}}, \qquad E_t = E_{t-1} \oplus Z_{L_t}.
  \end{aligned}
  \label{eq:cpi_message}
\end{equation}
Here $\oplus$ joins image-token blocks. The accumulated evidence counts the working view once and adds each crop. In $M_t$, both views share the offset $t_0+\Delta t_{\mathrm{seq}}$, which cancels in their relative M-RoPE rotation $\mathbf{R}(p_{L_t}-p_{\bar{G}})$. ResVision \textbf{maintains a target-patch similarity of $0.592$} as the intervening reasoning grows from 0 to 4,096 tokens (\cref{fig:mrope_cross_turn}).

\paragraph{3. Multi-Turn Reasoning Loop.}
The model can answer from the working view or inspect an ambiguous region. \texttt{crop} returns the paired observation $M_t$, and \texttt{read\_pixels} supplies textual RGB values when color is decisive. Each observation informs the next decision about where to look or whether to answer. The final prediction $\hat{y}=\operatorname{Decode}_{\theta}(q,H_t)$ uses the question $q$ and the conversation $H_t$ through the current turn.

\section{Experiments}
\label{sec:experiments}

We first ask what better sampling can achieve in a single pass, where resizing is the only intervention (\cref{sec:exp_direct}). We then test whether pairing the crop with the working image improves multi-turn reasoning and measure what each component contributes (\cref{sec:exp_agentic}).

\ifdefined\ispevekreport
\begin{wrapfigure}{r}{0.48\linewidth}
  \captionsetup{justification=raggedright,singlelinecheck=false}
  \centering
  \includegraphics[width=\linewidth]{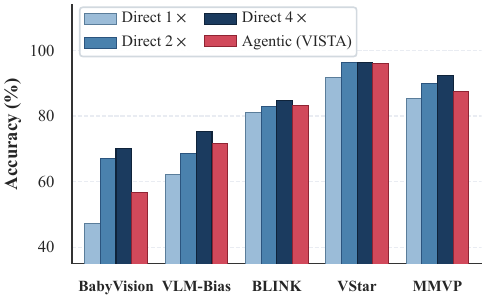}
  \caption{\textbf{Global upscaling on five fine-grained benchmarks.} \texttt{Qwen3.8-Flash-Next} compares Direct at $1\times$, $2\times$, and $4\times$ with \texttt{VISTA}.}
  \label{fig:five_bench_upscaling}
\end{wrapfigure}
\fi

\paragraph{Evaluation Setup.}
To see whether the gains carry across models, we run the main comparisons on Qwen3.8-Flash-Next, GLM-5.3-Flash, DeepSeek-V4.1-Flash, and Kimi-K3 in thinking mode. Qwen3.8 judges score the answers. Our broad comparison uses \textbf{MM400}, which spans 29 public multimodal benchmarks. We sample up to 400 test questions at random per benchmark and keep all questions from smaller ones, for a total of 11,221.

To see how the methods handle fine visual detail, we use a \textbf{five-benchmark suite} of 1,679 questions for the scale sweep and the controlled comparisons of inspection and token allocation. BabyVision, VLMsAreBiased, and BLINK come from MM400. VStarBench adds detail-rich scenes with small targets, and MMVP adds visually similar images that differ in subtle but decisive ways. Macro-29 and Macro-5 weight benchmarks equally, while Micro accuracy weights questions equally.

\subsection{Direct Inference with \SMR}
\label{sec:exp_direct}

\ifdefined\ispevekreport\else
\begin{figure}[htbp]
  \centering
  \includegraphics[width=0.58\linewidth]{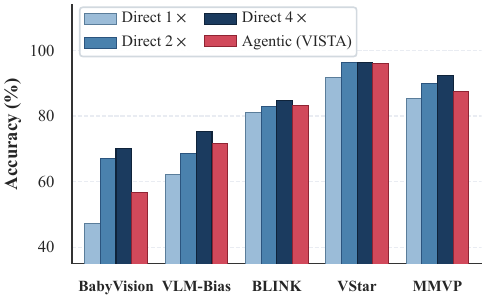}
  \caption{\textbf{Global upscaling on five fine-grained benchmarks.} \texttt{Qwen3.8-Flash-Next} compares Direct at $1\times$, $2\times$, and $4\times$ with \texttt{VISTA}.}
  \label{fig:five_bench_upscaling}
\end{figure}
\fi
\definecolor{scoregaincolor}{HTML}{D11D38}
\definecolor{scoredropcolor}{HTML}{215BDD}
\newcommand{\scoregain}[1]{{\scriptsize\textcolor{scoregaincolor}{(+#1)}}}
\newcommand{\scoredrop}[1]{{\scriptsize\textcolor{scoredropcolor}{(-#1)}}}
\newcommand{\scoretie}[1]{{\scriptsize\textcolor{black!55}{(+#1)}}}
\newcommand{\ablationdelta}[1]{\makebox[3.5em][r]{#1}}

\begin{table*}[t]
\centering
\caption{\textbf{Direct inference and \SMR across the 29 MM400 benchmarks.} Changes from native resolution are in percentage points.}
\label{tab:smart_resize_29}
\footnotesize
\setlength{\tabcolsep}{2.6pt}
\renewcommand{\arraystretch}{0.96}
\resizebox{\linewidth}{!}{%
\begin{tabular}{ll >{\columncolor{baselinegray}}cc >{\columncolor{baselinegray}}cc >{\columncolor{baselinegray}}cc >{\columncolor{baselinegray}}cc}
\toprule
& & \multicolumn{2}{c}{\textbf{Qwen3.8-Flash-Next}} & \multicolumn{2}{c}{\textbf{GLM-5.3-Flash}} & \multicolumn{2}{c}{\textbf{DeepSeek-V4.1-Flash}} & \multicolumn{2}{c}{\textbf{Kimi-K3}} \\
\cmidrule(lr){3-4} \cmidrule(lr){5-6} \cmidrule(lr){7-8} \cmidrule(lr){9-10}
\textbf{Domain} & \textbf{Benchmark} & \texttt{1x} & \SMR & \texttt{1x} & \SMR & \texttt{1x} & \SMR & \texttt{1x} & \SMR \\
\midrule
\multirow{5}{*}{\textit{Math \& Logic}}
& BabyVision & 47.16 & \textbf{70.88} \scoregain{23.7} & 40.46 & \textbf{50.61} \scoregain{10.1} & 33.51 & \textbf{57.47} \scoregain{24.0} & 55.07 & \textbf{74.05} \scoregain{19.0} \\
& VisuLogic & 44.75 & \textbf{49.75} \scoregain{5.0} & \textbf{47.00} & 42.86 \scoredrop{4.1} & 47.25 & \textbf{56.75} \scoregain{9.5} & 45.73 & \textbf{62.00} \scoregain{16.3} \\
& Kangaroo2025 & 88.50 & \textbf{92.00} \scoregain{3.5} & \textbf{86.75} & 85.28 \scoredrop{1.5} & \textbf{86.75} & 86.50 \scoredrop{0.3} & \textbf{94.00} & 93.25 \scoredrop{0.8} \\
& MathVision & 91.50 & \textbf{94.50} \scoregain{3.0} & 86.25 & \textbf{86.53} \scoregain{0.3} & 92.50 & \textbf{95.25} \scoregain{2.8} & 96.24 & \textbf{96.75} \scoregain{0.5} \\
& Rebus & \textbf{69.67} & 67.57 \scoredrop{2.1} & \textbf{73.87} & 72.26 \scoredrop{1.6} & 75.38 & \textbf{77.78} \scoregain{2.4} & 82.88 & \textbf{83.43} \scoregain{0.6} \\
\cmidrule(lr){1-10}
\multirow{4}{*}{\textit{Chart \& OCR}}
& OCR-Reasoning & 76.50 & \textbf{79.25} \scoregain{2.8} & \textbf{77.50} & 75.87 \scoredrop{1.6} & 75.50 & \textbf{77.25} \scoregain{1.8} & 78.25 & \textbf{79.00} \scoregain{0.8} \\
& CharXiv-RQ & 81.75 & \textbf{86.00} \scoregain{4.3} & 83.00 & \textbf{84.15} \scoregain{1.2} & 79.50 & \textbf{80.00} \scoregain{0.5} & 86.00 & \textbf{87.25} \scoregain{1.3} \\
& ChartMuseum & 81.00 & \textbf{83.50} \scoregain{2.5} & 79.75 & \textbf{80.59} \scoregain{0.8} & 77.25 & \textbf{77.75} \scoregain{0.5} & 81.50 & \textbf{85.25} \scoregain{3.8} \\
& CharXiv-Des & 95.50 & \textbf{96.25} \scoregain{0.8} & \textbf{95.00} & 94.78 \scoredrop{0.2} & 95.75 & \textbf{96.00} \scoregain{0.3} & \textbf{96.25} & 95.25 \scoredrop{1.0} \\
\cmidrule(lr){1-10}
\multirow{3}{*}{\textit{Halluc. \& Count}}
& VLMsAreBiased & 62.17 & \textbf{75.00} \scoregain{12.8} & 48.33 & \textbf{49.28} \scoregain{0.9} & 43.17 & \textbf{59.00} \scoregain{15.8} & 46.17 & \textbf{48.75} \scoregain{2.6} \\
& HallusionBench & 84.25 & \textbf{86.50} \scoregain{2.3} & 82.50 & \textbf{84.49} \scoregain{2.0} & 82.00 & \textbf{83.25} \scoregain{1.3} & \textbf{85.00} & 81.00 \scoredrop{4.0} \\
& CountQA & 72.25 & \textbf{73.50} \scoregain{1.3} & 59.50 & \textbf{60.20} \scoregain{0.7} & \textbf{60.50} & 57.25 \scoredrop{3.3} & \textbf{65.25} & 64.50 \scoredrop{0.8} \\
\cmidrule(lr){1-10}
\multirow{6}{*}{\textit{Embodied \& Spat.}}
& BLINK & 81.00 & \textbf{85.75} \scoregain{4.8} & 83.33 & \textbf{83.81} \scoregain{0.5} & 79.25 & \textbf{81.25} \scoregain{2.0} & 84.83 & \textbf{88.00} \scoregain{3.2} \\
& MMSI & 51.75 & \textbf{53.00} \scoregain{1.3} & 47.50 & \textbf{48.55} \scoregain{1.1} & 43.00 & \textbf{46.50} \scoregain{3.5} & \textbf{56.25} & 53.52 \scoredrop{2.7} \\
& ERQA & 68.25 & \textbf{69.50} \scoregain{1.3} & \textbf{69.25} & 65.35 \scoredrop{3.9} & 65.75 & \textbf{66.75} \scoregain{1.0} & 66.25 & \textbf{68.25} \scoregain{2.0} \\
& MindCube & 78.50 & \textbf{80.50} \scoregain{2.0} & \textbf{69.50} & 67.90 \scoredrop{1.6} & \textbf{80.75} & 80.00 \scoredrop{0.8} & \textbf{87.75} & 86.25 \scoredrop{1.5} \\
& CV-Bench & 90.75 & \textbf{92.25} \scoregain{1.5} & 90.75 & \textbf{91.78} \scoregain{1.0} & 88.50 & \textbf{89.00} \scoregain{0.5} & 91.75 & \textbf{92.25} \scoregain{0.5} \\
& OmniSpatial & \textbf{78.00} & 76.50 \scoredrop{1.5} & 67.75 & \textbf{68.01} \scoregain{0.3} & \textbf{75.00} & 74.50 \scoredrop{0.5} & 77.75 & \textbf{79.75} \scoregain{2.0} \\
\cmidrule(lr){1-10}
\multirow{4}{*}{\textit{Med. \& World}}
& MuirBench & 78.75 & \textbf{82.50} \scoregain{3.8} & 77.25 & \textbf{80.33} \scoregain{3.1} & 76.00 & \textbf{76.50} \scoregain{0.5} & \textbf{80.00} & 79.40 \scoredrop{0.6} \\
& MedXpertQA-MM & \textbf{71.00} & 69.50 \scoredrop{1.5} & 72.00 & \textbf{73.28} \scoregain{1.3} & 70.50 & \textbf{71.00} \scoregain{0.5} & \textbf{76.25} & 71.61 \scoredrop{4.6} \\
& WorldVQA & \textbf{39.50} & 38.25 \scoredrop{1.3} & 38.25 & \textbf{43.50} \scoregain{5.3} & 31.50 & \textbf{31.75} \scoregain{0.3} & \textbf{52.00} & 44.25 \scoredrop{7.8} \\
& RealWorldQA & \textbf{86.50} & \textbf{86.50} \scoretie{0.0} & \textbf{83.00} & 82.96 \scoredrop{0.0} & \textbf{82.50} & 82.25 \scoredrop{0.3} & \textbf{84.00} & \textbf{84.00} \scoretie{0.0} \\
\cmidrule(lr){1-10}
\multirow{7}{*}{\textit{General Visual}}
& SimpleVQA & 52.50 & \textbf{56.50} \scoregain{4.0} & 63.25 & \textbf{65.16} \scoregain{1.9} & 63.50 & \textbf{64.00} \scoregain{0.5} & \textbf{72.00} & 67.50 \scoredrop{4.5} \\
& MMMU & 84.50 & \textbf{85.75} \scoregain{1.3} & 83.50 & \textbf{84.66} \scoregain{1.2} & \textbf{87.25} & 86.50 \scoredrop{0.8} & \textbf{87.25} & 85.96 \scoredrop{1.3} \\
& MMMU-Pro & 80.25 & \textbf{81.50} \scoregain{1.3} & \textbf{78.75} & 77.90 \scoredrop{0.9} & 78.50 & \textbf{81.75} \scoregain{3.3} & 83.00 & \textbf{83.25} \scoregain{0.3} \\
& MMStar & \textbf{84.50} & 84.00 \scoredrop{0.5} & 82.25 & \textbf{83.63} \scoregain{1.4} & \textbf{83.00} & 81.00 \scoredrop{2.0} & 82.75 & \textbf{84.50} \scoregain{1.8} \\
& MMBench-CN & 94.00 & \textbf{94.25} \scoregain{0.3} & \textbf{91.50} & 90.59 \scoredrop{0.9} & \textbf{92.50} & 92.25 \scoredrop{0.3} & \textbf{93.00} & 92.75 \scoredrop{0.3} \\
& CCBench & 87.50 & \textbf{88.50} \scoregain{1.0} & 86.25 & \textbf{91.41} \scoregain{5.2} & \textbf{88.50} & 87.50 \scoredrop{1.0} & \textbf{91.50} & 88.50 \scoredrop{3.0} \\
& ZeroBench & \textbf{10.00} & \textbf{10.00} \scoretie{0.0} & 7.00 & \textbf{9.30} \scoregain{2.3} & 11.00 & \textbf{14.00} \scoregain{3.0} & \textbf{18.00} & 17.53 \scoredrop{0.5} \\
\midrule
\multicolumn{2}{l}{\textbf{Macro Average (29 Benchmarks)}} & 72.84 & \textbf{75.50} \scoregain{2.66} & 70.72 & \textbf{71.55} \scoregain{0.83} & 70.55 & \textbf{72.78} \scoregain{2.23} & 75.75 & \textbf{76.47} \scoregain{0.73} \\
\multicolumn{2}{l}{\textbf{Micro Accuracy (11,221 Items)}} & 74.56 & \textbf{77.30} \scoregain{2.74} & 72.44 & \textbf{73.45} \scoregain{1.01} & 72.16 & \textbf{74.34} \scoregain{2.18} & 77.27 & \textbf{78.01} \scoregain{0.74} \\
\bottomrule
\end{tabular}
}
\end{table*}

In the five-benchmark sweep, Direct accuracy improves as the image gets larger, with the biggest gain on BabyVision. At $4\times$, Qwen3.8 Direct reaches or exceeds VISTA on these tasks (\cref{fig:five_bench_upscaling}). We next test whether choosing the scale per image helps across MM400. We compare native resolution (\texttt{1x}) with \SMR at $p=32$~px/token, $f_c=0.77$~cycles/token, and a cap of 4,096 visual tokens. \textbf{\SMR improves Direct Macro-29 on all four models} by 0.73--2.66 points (\cref{tab:smart_resize_29}). BabyVision and VLMsAreBiased account for several of the largest gains, while the changes on general visual and world-knowledge tasks are more mixed. This variation motivates measuring the cost and accuracy of a fixed resize against a scale chosen per image.

\begin{table}[htbp]
\centering
\caption{\textbf{Ablation of \SMR policies and spectral percentiles on \texttt{Qwen3.8-Flash-Next} across MM400.} High-Freq covers Math, Chart/OCR, Hallucination, Counting, and Embodied. Low-Freq covers World Knowledge, Medical, and General. Changes from native resolution are in percentage points or tokens.}
\label{tab:ablation_smart_resize}
\small
\setlength{\tabcolsep}{5pt}
\renewcommand{\arraystretch}{1.12}
\begin{tabular*}{\linewidth}{@{\extracolsep{\fill}}lcccc@{}}
\toprule
\textbf{Resize Policy} & \textbf{High-Freq} & \textbf{Low-Freq} & \textbf{Macro-29} & \textbf{Avg. Prompt Tokens} \\
\midrule
\texttt{1x} (Native) & \cellcolor{baselinegray}74.63 & \cellcolor{baselinegray}69.91 & \cellcolor{baselinegray}72.84 & \cellcolor{baselinegray}1,389 \\
\texttt{Fixed-4k} & 78.40 \scoregain{3.77} & 69.70 \scoredrop{0.21} & 75.10 \scoregain{2.26} & 5,575 \scoregain{4,186} \\
\midrule
\SMR ($p_{50}$) & 75.57 \scoregain{0.94} & 69.84 \scoredrop{0.07} & 73.40 \scoregain{0.56} & 1,522 \scoregain{133} \\
\SMR ($p_{80}$) & 77.59 \scoregain{2.96} & 70.23 \scoregain{0.32} & 74.80 \scoregain{1.96} & 2,563 \scoregain{1,174} \\
\SMR ($p_{90}$) & \textbf{78.46} \scoregain{3.83} & \textbf{70.66} \scoregain{0.75} & \textbf{75.50} \scoregain{2.66} & \textbf{3,352} \scoregain{1,963} \\
\bottomrule
\end{tabular*}
\end{table}

\paragraph{Ablation on Resizing Policy and Spectral Percentiles.}
Under the same 4,096-visual-token cap, we compare \texttt{Fixed-4k} with grayscale FFT \SMR at $p_{50}$, $p_{80}$, and $p_{90}$ on Qwen3.8 (\cref{tab:ablation_smart_resize,tab:suppl_domain_ablation}). At $p_{90}$, \textbf{\SMR reaches $75.50\%$ Macro-29, $0.40$ points above Fixed-4k}. The two policies have similar high-frequency accuracy ($78.46\%$ and $78.40\%$), and \SMR scores $0.96$ points higher in low-frequency domains. In the resizing sweep, \SMR at $p_{90}$ uses 3,352 prompt tokens on average, $39.9\%$ fewer than Fixed-4k's 5,575.

\subsection{Agentic Perception Evaluation with ResVision}
\label{sec:exp_agentic}

\begin{table*}[t]
\centering
\caption{\textbf{Agentic visual reasoning across the 29 MM400 benchmarks.} Changes from VISTA are in percentage points.}
\label{tab:agentic_main}
\footnotesize
\setlength{\tabcolsep}{2.1pt}
\renewcommand{\arraystretch}{0.96}
\resizebox{\linewidth}{!}{%
\begin{tabular}{ll >{\columncolor{baselinegray}}cc >{\columncolor{baselinegray}}cc >{\columncolor{baselinegray}}cc >{\columncolor{baselinegray}}cc}
\toprule
& & \multicolumn{2}{c}{\textbf{Qwen3.8-Flash-Next}} & \multicolumn{2}{c}{\textbf{GLM-5.3-Flash}} & \multicolumn{2}{c}{\textbf{DeepSeek-V4.1-Flash}} & \multicolumn{2}{c}{\textbf{Kimi-K3}} \\
\cmidrule(lr){3-4} \cmidrule(lr){5-6} \cmidrule(lr){7-8} \cmidrule(lr){9-10}
\textbf{Domain} & \textbf{Benchmark} & \texttt{VISTA} & \texttt{ResVision} & \texttt{VISTA} & \texttt{ResVision} & \texttt{VISTA} & \texttt{ResVision} & \texttt{VISTA} & \texttt{ResVision} \\
\midrule
\multirow{5}{*}{\textit{Math \& Logic}}
& BabyVision & 56.61 & \textbf{77.75} \scoregain{21.1} & 51.63 & \textbf{68.21} \scoregain{16.6} & 42.61 & \textbf{59.36} \scoregain{16.8} & 59.37 & \textbf{80.76} \scoregain{21.4} \\
& VisuLogic & 48.67 & \textbf{61.00} \scoregain{12.3} & \textbf{48.33} & 44.50 \scoredrop{3.8} & 39.08 & \textbf{63.00} \scoregain{23.9} & 50.33 & \textbf{64.75} \scoregain{14.4} \\
& Kangaroo2025 & 91.67 & \textbf{92.25} \scoregain{0.6} & \textbf{91.50} & 87.00 \scoredrop{4.5} & 87.92 & \textbf{91.25} \scoregain{3.3} & 95.58 & \textbf{96.00} \scoregain{0.4} \\
& MathVision & 92.83 & \textbf{96.00} \scoregain{3.2} & 90.67 & \textbf{93.00} \scoregain{2.3} & 89.67 & \textbf{95.50} \scoregain{5.8} & 94.00 & \textbf{97.50} \scoregain{3.5} \\
& Rebus & 58.06 & \textbf{72.67} \scoregain{14.6} & 69.77 & \textbf{71.47} \scoregain{1.7} & 65.57 & \textbf{77.78} \scoregain{12.2} & \textbf{83.78} & 80.18 \scoredrop{3.6} \\
\cmidrule(lr){1-10}
\multirow{4}{*}{\textit{Chart \& OCR}}
& OCR-Reasoning & 77.25 & \textbf{81.50} \scoregain{4.3} & 76.33 & \textbf{79.25} \scoregain{2.9} & 76.17 & \textbf{79.25} \scoregain{3.1} & 79.83 & \textbf{85.00} \scoregain{5.2} \\
& CharXiv-RQ & 84.92 & \textbf{89.00} \scoregain{4.1} & \textbf{86.58} & 85.25 \scoredrop{1.3} & 79.92 & \textbf{85.25} \scoregain{5.3} & 89.83 & \textbf{91.00} \scoregain{1.2} \\
& ChartMuseum & 83.75 & \textbf{86.75} \scoregain{3.0} & 83.42 & \textbf{84.75} \scoregain{1.3} & 78.33 & \textbf{82.50} \scoregain{4.2} & 85.42 & \textbf{90.50} \scoregain{5.1} \\
& CharXiv-Des & 95.33 & \textbf{96.50} \scoregain{1.2} & \textbf{95.75} & 95.25 \scoredrop{0.5} & 95.83 & \textbf{96.25} \scoregain{0.4} & 96.00 & \textbf{96.50} \scoregain{0.5} \\
\cmidrule(lr){1-10}
\multirow{3}{*}{\textit{Halluc. \& Count}}
& VLMsAreBiased & 71.58 & \textbf{88.75} \scoregain{17.2} & 63.42 & \textbf{69.92} \scoregain{6.5} & 51.58 & \textbf{60.17} \scoregain{8.6} & 53.58 & \textbf{60.50} \scoregain{6.9} \\
& HallusionBench & 87.00 & \textbf{88.75} \scoregain{1.8} & \textbf{86.00} & 83.75 \scoredrop{2.3} & 87.08 & \textbf{88.25} \scoregain{1.2} & 85.58 & \textbf{86.25} \scoregain{0.7} \\
& CountQA & 74.08 & \textbf{74.50} \scoregain{0.4} & \textbf{64.00} & 61.25 \scoredrop{2.8} & 63.00 & \textbf{65.75} \scoregain{2.8} & \textbf{71.83} & 70.25 \scoredrop{1.6} \\
\cmidrule(lr){1-10}
\multirow{6}{*}{\textit{Embodied \& Spat.}}
& BLINK & 83.33 & \textbf{86.00} \scoregain{2.7} & 82.92 & \textbf{86.50} \scoregain{3.6} & 79.00 & \textbf{81.50} \scoregain{2.5} & 87.33 & \textbf{89.25} \scoregain{1.9} \\
& MMSI & 52.83 & \textbf{53.00} \scoregain{0.2} & 50.75 & \textbf{51.75} \scoregain{1.0} & 50.17 & \textbf{51.00} \scoregain{0.8} & \textbf{55.25} & 54.75 \scoredrop{0.5} \\
& ERQA & 69.33 & \textbf{75.50} \scoregain{6.2} & 68.75 & \textbf{75.00} \scoregain{6.3} & 65.33 & \textbf{71.50} \scoregain{6.2} & 69.67 & \textbf{73.50} \scoregain{3.8} \\
& MindCube & 84.58 & \textbf{85.75} \scoregain{1.2} & 81.17 & \textbf{81.50} \scoregain{0.3} & 60.92 & \textbf{80.50} \scoregain{19.6} & \textbf{87.75} & 87.50 \scoredrop{0.3} \\
& CV-Bench & 91.25 & \textbf{93.00} \scoregain{1.8} & 91.67 & \textbf{91.75} \scoregain{0.1} & \textbf{91.00} & 90.50 \scoredrop{0.5} & 93.25 & \textbf{94.00} \scoregain{0.8} \\
& OmniSpatial & \textbf{78.17} & 77.50 \scoredrop{0.7} & 73.08 & \textbf{74.50} \scoregain{1.4} & 71.83 & \textbf{75.00} \scoregain{3.2} & \textbf{77.08} & 73.50 \scoredrop{3.6} \\
\cmidrule(lr){1-10}
\multirow{4}{*}{\textit{Med. \& World}}
& MuirBench & 78.75 & \textbf{79.25} \scoregain{0.5} & \textbf{80.25} & 78.75 \scoredrop{1.5} & \textbf{74.67} & 73.75 \scoredrop{0.9} & 77.25 & \textbf{78.75} \scoregain{1.5} \\
& MedXpertQA-MM & \textbf{71.25} & 70.75 \scoredrop{0.5} & 72.75 & \textbf{75.50} \scoregain{2.8} & \textbf{69.83} & 69.50 \scoredrop{0.3} & \textbf{75.17} & 74.25 \scoredrop{0.9} \\
& WorldVQA & 30.75 & \textbf{39.00} \scoregain{8.3} & 41.25 & \textbf{47.00} \scoregain{5.8} & 31.33 & \textbf{36.50} \scoregain{5.2} & 46.92 & \textbf{51.50} \scoregain{4.6} \\
& RealWorldQA & 88.17 & \textbf{89.00} \scoregain{0.8} & 85.58 & \textbf{85.75} \scoregain{0.2} & 84.08 & \textbf{84.50} \scoregain{0.4} & 83.17 & \textbf{83.25} \scoregain{0.1} \\
\cmidrule(lr){1-10}
\multirow{7}{*}{\textit{General Visual}}
& SimpleVQA & 57.75 & \textbf{61.25} \scoregain{3.5} & 67.83 & \textbf{71.00} \scoregain{3.2} & 64.17 & \textbf{66.00} \scoregain{1.8} & \textbf{73.17} & 72.75 \scoredrop{0.4} \\
& MMMU & \textbf{84.83} & 84.50 \scoredrop{0.3} & \textbf{86.50} & 85.50 \scoredrop{1.0} & 85.42 & \textbf{86.25} \scoregain{0.8} & 85.75 & \textbf{88.25} \scoregain{2.5} \\
& MMMU-Pro & 78.58 & \textbf{80.50} \scoregain{1.9} & 81.00 & \textbf{81.25} \scoregain{0.3} & 79.33 & \textbf{81.50} \scoregain{2.2} & 82.50 & \textbf{85.50} \scoregain{3.0} \\
& MMStar & 82.92 & \textbf{85.25} \scoregain{2.3} & 82.08 & \textbf{84.00} \scoregain{1.9} & 80.08 & \textbf{81.00} \scoregain{0.9} & 81.67 & \textbf{82.75} \scoregain{1.1} \\
& MMBench-CN & \textbf{94.17} & 92.75 \scoredrop{1.4} & 93.75 & \textbf{94.00} \scoregain{0.3} & 93.17 & \textbf{93.75} \scoregain{0.6} & \textbf{93.08} & 93.00 \scoredrop{0.1} \\
& CCBench & 86.75 & \textbf{89.00} \scoregain{2.3} & 87.25 & \textbf{89.75} \scoregain{2.5} & 87.17 & \textbf{89.00} \scoregain{1.8} & 91.92 & \textbf{93.25} \scoregain{1.3} \\
& ZeroBench & 9.67 & \textbf{10.00} \scoregain{0.3} & 10.67 & \textbf{11.00} \scoregain{0.3} & 8.67 & \textbf{17.00} \scoregain{8.3} & 15.33 & \textbf{22.00} \scoregain{6.7} \\
\midrule
\multicolumn{2}{l}{\textbf{Macro Average (29 Benchmarks)}} & 73.96 & \textbf{77.84} \scoregain{3.88} & 73.95 & \textbf{75.45} \scoregain{1.50} & 70.10 & \textbf{74.93} \scoregain{4.83} & 76.60 & \textbf{79.20} \scoregain{2.60} \\
\multicolumn{2}{l}{\textbf{Micro Accuracy (11,221 Items)}} & 75.79 & \textbf{79.69} \scoregain{3.89} & 75.69 & \textbf{77.21} \scoregain{1.51} & 71.80 & \textbf{76.48} \scoregain{4.68} & 78.21 & \textbf{80.73} \scoregain{2.51} \\
\bottomrule
\end{tabular}
}
\end{table*}

\paragraph{Benchmarks and Protocol.}
We compare ResVision with VISTA on MM400 to see how well each uses a closer view across tasks (\cref{tab:agentic_main}). On the five-benchmark suite\ifdefined\ispevekreport{} (\cref{sec:supp_five_benchmarks})\fi, both methods can make up to four tool calls per question.

\paragraph{Main Agentic Results.}
On MM400, \textbf{ResVision improves on the existing VISTA baseline on all four models}, with Macro-29 gains of 1.50--4.83 points. Kimi-K3 reaches the highest Macro-29 at 79.20, and DeepSeek shows the largest difference from VISTA at 4.83 points. In the three-run five-benchmark comparison (\cref{tab:smr_cpi_ablation}), ResVision scores 6.39 Macro-5 points above VISTA on Kimi-K3 and 9.02 on Qwen3.8.
\begin{itemize}
  \item \textbf{Perception and grounding tasks.} All four models score higher on \texttt{BabyVision} (16.6--21.4 points), \texttt{VLMsAreBiased} (6.5--17.2), and \texttt{ChartMuseum} (1.3--5.1). On \texttt{VisuLogic}, the difference ranges from $-3.8$ points for GLM to $+23.9$ for DeepSeek. These tasks require both fine local detail and a global reference such as a candidate set, axis, or table header.
  \item \textbf{Mixed changes on general visual tasks.} On \texttt{SimpleVQA}, for example, ResVision improves Qwen by 3.5 points but lowers Kimi-K3 by 0.4 points. The strongest gains are concentrated where the answer turns on a visual distinction.
\end{itemize}

\begin{table}[htbp]
\centering
\caption{\textbf{Macro-29 component comparison of \SMR and CPI (\%).} Changes are relative to Agentic Base.}
\label{tab:macro29_cpi_only}
\begingroup
\setlength{\tabcolsep}{3.6pt}
\renewcommand{\arraystretch}{1.16}
\small
\begin{tabular*}{\linewidth}{@{\extracolsep{\fill}}lrrrr@{}}
\toprule
\textbf{Model} & Agentic Base & +SMR & +CPI & +SMR+CPI \\
\midrule
Qwen3.8-Flash-Next & \cellcolor{baselinegray}73.96 & 75.96 \scoregain{2.00} & 75.06 \scoregain{1.10} & \textbf{77.84} \scoregain{3.88} \\
Kimi-K3 & \cellcolor{baselinegray}76.60 & 78.27 \scoregain{1.67} & 78.57 \scoregain{1.97} & \textbf{79.20} \scoregain{2.60} \\
DeepSeek-V4.1-Flash & \cellcolor{baselinegray}70.10 & 70.87 \scoregain{0.77} & 71.57 \scoregain{1.47} & \textbf{74.93} \scoregain{4.83} \\
GLM-5.3-Flash & \cellcolor{baselinegray}73.95 & 75.30 \scoregain{1.35} & 75.09 \scoregain{1.14} & \textbf{75.45} \scoregain{1.50} \\
\bottomrule
\end{tabular*}
\endgroup

\end{table}

\paragraph{Ablation of \SMR, CPI, and Adaptive Allocation.}
To separate the benefit of a better initial image from the benefit of joint inspection, we vary \SMR and CPI independently (\cref{tab:smr_cpi_ablation,tab:macro29_cpi_only}). On Macro-5, adding \SMR to Direct improves Kimi-K3 by 5.26 points and Qwen3.8 by 8.88 points. On Macro-29, \SMR alone and CPI alone raise the four-model mean from $73.65\%$ to $75.10\%$ and $75.07\%$, respectively. \textbf{Full ResVision has the highest Macro-29 on every model}, averaging $76.86\%$.

We next hold the inspection policy fixed and vary how the working image is resized (\cref{tab:fixed_token_resize_ablation}). With CPI and a fixed allocation, 2,048 tokens gives the best Macro-5 for both Kimi-K3 and Qwen3.8. Adaptive allocation with CPI improves on that setting by 3.31 and 3.36 points, respectively. It also improves on the SMR-only control (Original) by 2.21 and 1.85 points. The two comparisons show how choosing a scale per image and preserving context during inspection each contribute to the final result.

\begin{table*}[htbp]
\centering
\begin{minipage}[t]{0.495\linewidth}
  \captionsetup{font=footnotesize,skip=3pt}
  \captionof{table}{\textbf{Component ablations.} Macro-5 mean $\pm$ SD over three runs. Changes from each Base.}
  \label{tab:smr_cpi_ablation}
  \centering
\fontsize{7.8}{9.4}\selectfont
\setlength{\tabcolsep}{2.0pt}
\renewcommand{\arraystretch}{1.08}
\begin{tabular*}{\linewidth}{@{\extracolsep{\fill}}lrr@{}}
\toprule
Configuration & Kimi-K3 & Qwen 3.8 \\
\midrule
\rowcolor{ablmodelband}\multicolumn{3}{l}{\textbf{Direct}} \\
Base & \cellcolor{baselinegray}72.30\,\textpm\,0.40 \ablationdelta{} & \cellcolor{baselinegray}73.47\,\textpm\,0.29 \ablationdelta{} \\
+SMR & 77.57\,\textpm\,0.55 \ablationdelta{\scoregain{5.26}} & 82.35\,\textpm\,0.96 \ablationdelta{\scoregain{8.88}} \\
\addlinespace[2pt]
\rowcolor{ablmodelband}\multicolumn{3}{l}{\textbf{Agentic}} \\
Base & \cellcolor{baselinegray}75.64\,\textpm\,0.74 \ablationdelta{} & \cellcolor{baselinegray}78.99\,\textpm\,0.80 \ablationdelta{} \\
+CPI & 79.79\,\textpm\,0.97 \ablationdelta{\scoregain{4.15}} & 84.38\,\textpm\,0.49 \ablationdelta{\scoregain{5.39}} \\
+SMR & 79.83\,\textpm\,0.44 \ablationdelta{\scoregain{4.19}} & 86.16\,\textpm\,0.32 \ablationdelta{\scoregain{7.17}} \\
\rowcolor{ablglobal!8}
\textbf{+SMR+CPI} & \textbf{82.03}\,\textpm\,0.73 \ablationdelta{\scoregain{6.39}} & \textbf{88.01}\,\textpm\,0.56 \ablationdelta{\scoregain{9.02}} \\
\bottomrule
\end{tabular*}

\end{minipage}\hfill
\begin{minipage}[t]{0.485\linewidth}
  \captionsetup{font=footnotesize,skip=3pt}
  \captionof{table}{\textbf{Adaptive versus fixed allocation.} Macro-5 mean $\pm$ SD over three runs. Changes from Original.}
  \label{tab:fixed_token_resize_ablation}
  \centering
\fontsize{7.8}{9.4}\selectfont
\setlength{\tabcolsep}{2.0pt}
\renewcommand{\arraystretch}{1.08}
\begin{tabular*}{\linewidth}{@{\extracolsep{\fill}}lrr@{}}
\toprule
Allocation & Kimi-K3 & Qwen 3.8 \\
\midrule
Original & \cellcolor{baselinegray}79.83\,\textpm\,0.44 \ablationdelta{} & \cellcolor{baselinegray}86.16\,\textpm\,0.32 \ablationdelta{} \\
Fixed 1,024 & 78.24\,\textpm\,0.45 \ablationdelta{\scoredrop{1.59}} & 82.64\,\textpm\,0.79 \ablationdelta{\scoredrop{3.52}} \\
Fixed 2,048 & 78.72\,\textpm\,0.37 \ablationdelta{\scoredrop{1.10}} & 84.65\,\textpm\,0.97 \ablationdelta{\scoredrop{1.52}} \\
Fixed 3,072 & 76.43\,\textpm\,0.70 \ablationdelta{\scoredrop{3.40}} & 83.37\,\textpm\,0.38 \ablationdelta{\scoredrop{2.79}} \\
Fixed 4,096 & 74.26\,\textpm\,1.60 \ablationdelta{\scoredrop{5.56}} & 84.13\,\textpm\,0.78 \ablationdelta{\scoredrop{2.04}} \\
Fixed 5,120 & 73.58\,\textpm\,0.24 \ablationdelta{\scoredrop{6.24}} & 83.29\,\textpm\,0.56 \ablationdelta{\scoredrop{2.88}} \\
\rowcolor{ablglobal!8}
\textbf{SMR (adaptive)} & \textbf{82.03}\,\textpm\,0.73 \ablationdelta{\scoregain{2.21}} & \textbf{88.01}\,\textpm\,0.56 \ablationdelta{\scoregain{1.85}} \\
\bottomrule
\end{tabular*}

\end{minipage}
\end{table*}

\paragraph{Inference-Configuration Sensitivity.}
With the visual pipeline fixed, we vary the decoding settings on BabyVision (\cref{tab:configuration_robustness}). Reasoning effort has the largest effect. Moving from Low to Max raises accuracy by 22.68 points on Kimi-K3 and 8.68 on Qwen3.8. The tested temperatures and generation ceilings change accuracy by at most 3.01 points relative to the defaults.

\paragraph{End-to-End Latency.}
We measure the time cost of the resulting inspection loop on the five-benchmark suite (\cref{tab:latency}). ResVision takes $1.07$--$1.36\times$ as long as Direct. Relative to the other multi-turn method, VISTA, it is 2.8\% faster on Qwen3.8 and 3.5--9.7\% slower on the other three models.

\begin{table*}[htbp]
\centering
\begin{minipage}[t]{0.535\linewidth}
  \captionsetup{font=footnotesize,skip=3pt}
  \captionof{table}{\textbf{Configuration sensitivity on BabyVision (\%).} Changes from temperature 0, Max effort, and 131K tokens.}
  \label{tab:configuration_robustness}
  \centering
\fontsize{7.5}{9.0}\selectfont
\setlength{\tabcolsep}{1.8pt}
\renewcommand{\arraystretch}{1.06}
\begin{tabular*}{\linewidth}{@{\extracolsep{\fill}}llrr@{}}
\toprule
Factor & Setting & Kimi-K3 & Qwen 3.8 \\
\midrule
Temperature & \cellcolor{baselinegray}0 (default) & \cellcolor{baselinegray}\textbf{80.76}\,\textpm\,1.07 \ablationdelta{} & \cellcolor{baselinegray}77.75\,\textpm\,1.05 \ablationdelta{} \\
 & 0.5 & 80.50\,\textpm\,0.30 \ablationdelta{\scoredrop{0.26}} & \textbf{79.38}\,\textpm\,1.57 \ablationdelta{\scoregain{1.63}} \\
 & 1.0 & 77.75\,\textpm\,0.54 \ablationdelta{\scoredrop{3.01}} & 76.98\,\textpm\,0.54 \ablationdelta{\scoredrop{0.77}} \\
\addlinespace[2pt]
Effort & Low & 58.08\,\textpm\,2.84 \ablationdelta{\scoredrop{22.68}} & 69.07\,\textpm\,0.93 \ablationdelta{\scoredrop{8.68}} \\
 & Medium & 74.74\,\textpm\,2.30 \ablationdelta{\scoredrop{6.02}} & 75.86\,\textpm\,1.55 \ablationdelta{\scoredrop{1.89}} \\
 & \cellcolor{baselinegray}Max (default) & \cellcolor{baselinegray}\textbf{80.76}\,\textpm\,1.07 \ablationdelta{} & \cellcolor{baselinegray}\textbf{77.75}\,\textpm\,1.05 \ablationdelta{} \\
\addlinespace[2pt]
Token cap & 32K & 81.10\,\textpm\,0.74 \ablationdelta{\scoregain{0.34}} & \textbf{78.70}\,\textpm\,1.30 \ablationdelta{\scoregain{0.95}} \\
 & 65K & \textbf{81.27}\,\textpm\,0.39 \ablationdelta{\scoregain{0.51}} & 76.38\,\textpm\,1.16 \ablationdelta{\scoredrop{1.37}} \\
 & \cellcolor{baselinegray}131K (default) & \cellcolor{baselinegray}80.76\,\textpm\,1.07 \ablationdelta{} & \cellcolor{baselinegray}77.75\,\textpm\,1.05 \ablationdelta{} \\
\bottomrule
\end{tabular*}

\end{minipage}\hfill
\begin{minipage}[t]{0.445\linewidth}
  \captionsetup{font=footnotesize,skip=3pt}
  \captionof{table}{\textbf{End-to-end latency (s).} Ratios to Direct appear in parentheses.}
  \label{tab:latency}
  \begingroup
\centering
\fontsize{7.7}{9.2}\selectfont
\setlength{\tabcolsep}{1.8pt}
\renewcommand{\arraystretch}{1.14}
\newcommand{\latencyratio}[1]{{\scriptsize\textcolor{black}{($#1\times$)}}}
\newcommand{\latencycell}[2]{#1\enspace\latencyratio{#2}}
\begin{tabular*}{\linewidth}{@{\extracolsep{\fill}}lrrr@{}}
\toprule
Model & Direct & VISTA & ResVision \\
\midrule
Kimi-K3 & \cellcolor{baselinegray}\latencycell{248.5}{1.00} & \latencycell{264.8}{1.07} & \latencycell{274.2}{1.10} \\
Qwen & \cellcolor{baselinegray}\latencycell{51.5}{1.00} & \latencycell{68.5}{1.33} & \latencycell{66.6}{1.29} \\
GLM & \cellcolor{baselinegray}\latencycell{193.6}{1.00} & \latencycell{240.7}{1.24} & \latencycell{263.6}{1.36} \\
DeepSeek & \cellcolor{baselinegray}\latencycell{100.4}{1.00} & \latencycell{97.7}{0.97} & \latencycell{107.2}{1.07} \\
\midrule
Mean ratio & \cellcolor{baselinegray}\latencyratio{1.00} & \latencyratio{1.15} & \latencyratio{1.20} \\
\bottomrule
\end{tabular*}
\endgroup

\end{minipage}
\end{table*}

\section{Related Work}
\label{sec:related}

\subsection{Visual Scaling and Token Allocation}
A smooth photograph and a dense chart can have the same dimensions. In the chart, a tick mark that occupies a fraction of a visual token can determine the answer. Several designs make such details easier to access. NaViT packs inputs of different resolutions and aspect ratios within a token budget, Pix2Struct respects a patch budget while preserving aspect ratio, and FlexiViT supports multiple patch sizes within one model~\cite{dehghani2023navit,lee2023pix2struct,beyer2023flexivit}. Monkey and LLaVA-UHD use local tiles or variable-sized slices, and CogAgent adds a high-resolution branch~\cite{li2024monkey,xu2024llavauhd,hong2024cogagent}. TokenLearner, DynamicViT, and Token Merging adjust the token set inside the encoder by learning, pruning, or merging representations~\cite{ryoo2021tokenlearner,rao2021dynamicvit,bolya2023tome}. Motivated by the frequency sensitivity of visual representations~\cite{park2022vision}, \SMR uses the image spectrum and the encoder cutoff to choose the working resolution. This gives the chart more sampling when its fine structure calls for it, within the same token budget.

\subsection{Agentic Visual Reasoning}
A close-up can make a mark legible while hiding the row, axis, or object that gives it meaning. Active perception and recurrent visual attention study how a model directs its observations as a task unfolds~\cite{bajcsy1988active,mnih2014attention}, and language-model agents interleave observations with reasoning and tool calls~\cite{yao2023react}. VisProg, ViperGPT, MM-REACT, and LLaVA-Plus compose visual modules~\cite{gupta2023visprog,suris2023vipergpt,yang2023mmreact,rasheed2024llavaplus}. V* searches high-resolution images for small regions, VISTA retrieves local observations during generation, and VisualToolBench evaluates reasoning with visual tools~\cite{wu2024vstar,han2026vista,guo2025visualtoolbench}. ViP-LLaVA, Kosmos-2, and Ferret link language to image regions~\cite{cai2024vipllava,peng2024kosmos2,you2024ferret}, while Visual CoT, Interleaved CoT, and Sketchpad bring local views or visual intermediate steps into reasoning~\cite{shao2024visualcot,gao2025icot,hu2024sketchpad}. Visual Funnel combines focal detail and broader context through model-side attention~\cite{jung2026visualfunnel}. \textbf{ResVision brings both views together in the tool reply.} \SMR chooses a working scale from the image spectrum, and CPI returns that same working image with each enlarged crop. The model can read the mark and see which row, axis, or object it belongs to, without retraining the model.

\section{Conclusion}
\label{sec:conclusion}

How a visual agent samples and revisits an image shapes the evidence it can reason from. Our probes trace this dependence from individual tokens to the token grid and across inspection turns. \textbf{ResVision makes visual sampling and inspection part of inference.} SMR chooses a working scale from the image spectrum, and CPI keeps each enlarged crop linked to the full scene. Across four models on MM400, SMR raises Direct Macro-29 by 0.73--2.66 points, and ResVision gains 1.50--4.83 points over the existing VISTA baseline. These gains come from giving existing VLMs better access to visual evidence with the model weights fixed.

Acquiring visual evidence is a core problem for multimodal reasoning. Potential uses span reading dense charts and scanned documents, examining fine structures in microscopy images, locating defects on manufactured parts, and finding targets in cluttered scenes. \textbf{ResVision advances a guiding principle for multimodal intelligence by treating perception and reasoning as a feedback loop in which inference directs observation and new evidence revises inference.}

\section*{Contributors}
{\bfseries\noindent Guojian Zhan\textsuperscript{1,2,*},~Jiahao Xu\textsuperscript{1,\dag},~Yahui Liu\textsuperscript{1,\dag},~Yao Hu\textsuperscript{1},~Chuan Mu\textsuperscript{1}\par}
\vspace{0.25em}
{\small
  \noindent \textsuperscript{1}\,AllSpark Post-Training Team \quad \textsuperscript{2}\,Tsinghua University\par
  \noindent\textsuperscript{*}\,Work done during an internship at Xiaohongshu.\par
  \noindent\textsuperscript{\dag}\,Corresponding authors.\par
}

\clearpage
{\small
  \bibliographystyle{styles/ieeenat_fullname}
  \bibliography{main}
}

\clearpage
\ifdefined\ispevekreport\else
\setcounter{page}{1}
\fi
\maketitlesupplementary

\section{Spectral Estimation Details and Additional \SMR Ablations}
\label{sec:suppl_ablation}

The resize rule depends on how we measure the image spectrum. We describe the estimator and compare grayscale and edge-weighted spectra to show how that choice affects the token budget.

\paragraph{Estimating the grayscale spectrum.}
For an input image of width $W$ and height $H$, our client-side \SMR implementation (\cref{sec:method_direct}) converts the image to normalized grayscale, $I \in [0,1]^{H\times W}$. It divides the image into $64\times64$~px blocks and retains the $25\%$ with the highest spatial variance to discount blank margins. For each retained block, it removes the mean and applies a 2D Hann window before taking the 2D FFT. The resulting power is accumulated into 32 radial bins $P(f)$ spanning $f\in(0,0.5]$~cycles/px.

\paragraph{Choosing the resize scale.}
We extract the frequency $f_\eta$ at which the cumulative power reaches fraction $\eta$. The target scale is
\[
s_{\mathrm{spec}} =
  \operatorname{clip}\!\left(\frac{p f_\eta}{f_c},\,s_{\min},\,s_{\max}\right).
\]
We test $\eta=0.50$, $0.80$, and $0.90$. Here $p=32$~px/token, $f_c=0.77$~cycles/token, and the visual-token budget is $N_{\max}=4096$.

\paragraph{What Sobel pre-filtering changes.}
To test whether emphasizing edges changes how \SMR allocates tokens, we apply a $3\times3$ Sobel magnitude pre-filter before estimating the spectrum,
\[
g=\sqrt{g_x^2+g_y^2}.
\]
First-order differentiation weights power by $4\pi^2 f^2$, shifting cumulative energy toward higher frequencies. Taking the gradient magnitude also introduces second-harmonic components at step edges.

\paragraph{Accuracy and token cost.}
This shift toward higher frequencies pushes the target scale up. Across the 11,221 MM400 items, Sobel-FFT at $p_{90}$ sends $99.4\%$ of images to the 4,096-token cap. It averages 5,530 prompt tokens and scores $75.12\%$ Macro-29 and $76.94\%$ Micro, close to \texttt{Fixed-4k} ($75.10\%$ Macro-29 and 5,575 prompt tokens).

Grayscale FFT at $p_{90}$ preserves the $1/f^2$ spectral decay of natural scenes. In the resizing sweep, it keeps $53.4\%$ of images (5,992 of 11,221) below the cap, with a median target of 3,555 visual tokens. It uses 3,352 prompt tokens on average, $39.9\%$ fewer than \texttt{Fixed-4k}. At the main 4,096-token setting, it scores $75.50\%$ Macro-29 and $77.30\%$ Micro (\cref{tab:smart_resize_29}). The 4,608-token run scores $75.27\%$ and $77.11\%$, respectively, at 3,561 average prompt tokens.

\begin{table*}[h]
\centering
\caption{\textbf{Per-domain \SMR policy ablation on \texttt{Qwen3.8-Flash-Next}} (\cref{tab:ablation_smart_resize}). Changes from native resolution are in percentage points.}
\label{tab:suppl_domain_ablation}
\footnotesize
\setlength{\tabcolsep}{1.45pt}
\renewcommand{\arraystretch}{1.0}
\begin{tabular}{lcc >{\columncolor{baselinegray}}c ccccc}
\toprule
\textbf{Capability Domain} & \textbf{\# Benches} & \textbf{\# Items} & \textbf{\texttt{1x}} & \textbf{\texttt{Fixed-4k}} & \textbf{Sobel ($p_{90}$)} & \textbf{\SMR ($p_{50}$)} & \textbf{\SMR ($p_{80}$)} & \textbf{\SMR ($p_{90}$)} \\
\midrule
Hallucination \& Bias & 2 & 800 & 73.21 & 80.12 {\scriptsize(+6.91)} & 80.12 {\scriptsize(+6.91)} & 74.88 {\scriptsize(+1.67)} & 79.12 {\scriptsize(+5.91)} & \textbf{80.75} {\scriptsize(+7.54)} \\
Math \& Visual Logic & 5 & 1,921 & 68.32 & 74.23 {\scriptsize(+5.91)} & 74.29 {\scriptsize(+5.97)} & 70.10 {\scriptsize(+1.78)} & 72.17 {\scriptsize(+3.85)} & \textbf{74.94} {\scriptsize(+6.62)} \\
Chart \& Document OCR & 4 & 1,600 & 83.69 & \textbf{86.50} {\scriptsize(+2.81)} & \textbf{86.50} {\scriptsize(+2.81)} & 85.19 {\scriptsize(+1.50)} & 86.12 {\scriptsize(+2.43)} & 86.25 {\scriptsize(+2.56)} \\
Counting & 1 & 400 & 72.25 & 72.50 {\scriptsize(+0.25)} & 72.50 {\scriptsize(+0.25)} & 72.25 {\scriptsize(+0.00)} & 73.25 {\scriptsize(+1.00)} & \textbf{73.50} {\scriptsize(+1.25)} \\
Embodied \& Spatial & 6 & 2,400 & 74.71 & \textbf{76.88} {\scriptsize(+2.17)} & \textbf{76.88} {\scriptsize(+2.17)} & 74.50 {\scriptsize(-0.21)} & 76.62 {\scriptsize(+1.91)} & 76.25 {\scriptsize(+1.54)} \\
\cmidrule(lr){1-9}
World Knowledge & 3 & 1,200 & 68.25 & 67.08 {\scriptsize(-1.17)} & 67.08 {\scriptsize(-1.17)} & 68.25 {\scriptsize(+0.00)} & 68.58 {\scriptsize(+0.33)} & \textbf{69.08} {\scriptsize(+0.83)} \\
Medical Imaging & 1 & 400 & 71.00 & 68.50 {\scriptsize(-2.50)} & 68.50 {\scriptsize(-2.50)} & \textbf{72.25} {\scriptsize(+1.25)} & 69.25 {\scriptsize(-1.75)} & 69.50 {\scriptsize(-1.50)} \\
General Visual & 7 & 2,500 & 70.46 & 71.00 {\scriptsize(+0.54)} & 71.04 {\scriptsize(+0.58)} & 70.18 {\scriptsize(-0.28)} & 71.07 {\scriptsize(+0.61)} & \textbf{71.50} {\scriptsize(+1.04)} \\
\midrule
\textbf{Macro Average (29 Benches)} & 29 & 11,221 & 72.84 & 75.10 {\scriptsize(+2.26)} & 75.12 {\scriptsize(+2.28)} & 73.40 {\scriptsize(+0.56)} & 74.80 {\scriptsize(+1.96)} & \textbf{75.50} {\scriptsize(+2.66)} \\
\textbf{Micro Accuracy (All Items)} & 29 & 11,221 & 74.56 & 76.92 {\scriptsize(+2.36)} & 76.94 {\scriptsize(+2.38)} & 75.14 {\scriptsize(+0.58)} & 76.63 {\scriptsize(+2.07)} & \textbf{77.30} {\scriptsize(+2.74)} \\
\textbf{Average Prompt Tokens} & 29 & 11,221 & 1,389 & 5,575 & 5,530 & 1,522 & 2,563 & \textbf{3,352} \\
\bottomrule
\end{tabular}
\end{table*}

\ifdefined\ispevekreport
\clearpage
\section{Benchmark Suites}
\label{sec:supp_eval_card}

\subsection{MM400 Benchmark Suite (29 Benchmarks)}
\label{sec:supp_mm400_benchmarks}

The main comparisons cover six domains, from chart reading and spatial reasoning to medical and general visual tasks (\cref{tab:smart_resize_29,tab:agentic_main}). MM400 combines $11{,}221$ items from 29 public benchmarks (\cref{tab:supp_mm400_benchmarks}). We sample up to 400 test items per benchmark and include the full BabyVision, Rebus, and ZeroBench snapshots. Macro-29 weights benchmarks equally, while Micro weights items equally.

\begin{suppTable}
\centering
\footnotesize
\setlength{\tabcolsep}{3pt}
\renewcommand{\arraystretch}{1.04}
\caption{\textbf{MM400 benchmark inventory.}}
\label{tab:supp_mm400_benchmarks}
\begin{tabular*}{\linewidth}{@{\extracolsep{\fill}}lr>{\raggedright\arraybackslash}p{0.57\linewidth}@{}}
\toprule
\textbf{Benchmark} & \textbf{Items} & \textbf{Primary task} \\
\midrule
\rowcolor{ablmodelband}\multicolumn{3}{@{}l}{\textbf{Math and Logic}} \\
\href{https://arxiv.org/abs/2601.06521}{BabyVision} & 388 & Discrimination, tracking, patterns \\
\href{https://visulogic-benchmark.github.io/VisuLogic/}{VisuLogic} & 400 & Shapes, transforms, and relations \\
\href{https://matharena.ai/kangaroo/}{Kangaroo2025} & 400 & Competition math with diagrams \\
\href{https://mathllm.github.io/mathvision/}{MathVision} & 400 & Image-grounded math problems \\
\href{https://arxiv.org/abs/2401.05604}{Rebus} & 333 & Picture and symbol wordplay \\
\rowcolor{ablmodelband}\multicolumn{3}{@{}l}{\textbf{Chart and OCR}} \\
\href{https://arxiv.org/abs/2505.17163}{OCR-Reasoning} & 400 & Reasoning over text-rich images \\
\href{https://charxiv.github.io/}{CharXiv-RQ} & 400 & Scientific chart reasoning \\
\href{https://chartmuseum-leaderboard.github.io/}{ChartMuseum} & 400 & QA over real-world charts \\
\href{https://charxiv.github.io/}{CharXiv-Des} & 400 & Axes, labels, and legends \\
\rowcolor{ablmodelband}\multicolumn{3}{@{}l}{\textbf{Hallucination and Counting}} \\
\href{https://github.com/anvo25/vlms-are-biased}{VLMsAreBiased} & 400 & Counterfactuals that test priors \\
\href{https://arxiv.org/abs/2310.14566}{HallusionBench} & 400 & Visual illusions and hallucination \\
\href{https://arxiv.org/abs/2508.06585}{CountQA} & 400 & Object counting in clutter \\
\rowcolor{ablmodelband}\multicolumn{3}{@{}l}{\textbf{Embodied and Spatial}} \\
\href{https://zeyofu.github.io/blink/}{BLINK} & 400 & Correspondence, depth, localization \\
\href{https://github.com/InternRobotics/MMSI-Bench}{MMSI} & 400 & Multi-image spatial inference \\
\href{https://github.com/embodiedreasoning/ERQA}{ERQA} & 400 & Embodied reasoning and actions \\
\href{https://openreview.net/pdf/f893f9f2c20e6ca9961483d8c4846a6c98355489.pdf}{MindCube} & 400 & Spatial models from limited views \\
\href{https://huggingface.co/datasets/nyu-visionx/CV-Bench}{CV-Bench} & 400 & 2D relations and 3D depth \\
\href{https://arxiv.org/abs/2506.03135}{OmniSpatial} & 400 & Spatial reasoning across viewpoints \\
\rowcolor{ablmodelband}\multicolumn{3}{@{}l}{\textbf{Medical and World Knowledge}} \\
\href{https://arxiv.org/abs/2406.09411}{MuirBench} & 400 & Comparing evidence across images \\
\href{https://arxiv.org/abs/2501.18362}{MedXpertQA-MM} & 400 & Expert multimodal medical QA \\
\href{https://arxiv.org/abs/2602.02537}{WorldVQA} & 400 & Image-grounded world knowledge \\
\href{https://huggingface.co/datasets/xai-org/RealworldQA}{RealWorldQA} & 400 & Spatial QA in real scenes \\
\rowcolor{ablmodelband}\multicolumn{3}{@{}l}{\textbf{General Visual}} \\
\href{https://openaccess.thecvf.com/content/ICCV2025/html/Cheng_SimpleVQA_Multimodal_Factuality_Evaluation_for_Multimodal_Large_Language_Models_ICCV_2025_paper.html}{SimpleVQA} & 400 & Visual factuality in English and Chinese \\
\href{https://arxiv.org/abs/2311.16502}{MMMU} & 400 & Multidisciplinary multimodal QA \\
\href{https://arxiv.org/abs/2409.02813}{MMMU-Pro} & 400 & Harder multidisciplinary QA \\
\href{https://github.com/MMStar-Benchmark/MMStar}{MMStar} & 400 & Visually grounded multimodal QA \\
\href{https://github.com/open-compass/MMBench}{MMBench-CN} & 400 & Chinese visual perception and reasoning \\
\href{https://github.com/open-compass/MMBench}{CCBench} & 400 & Chinese cultural knowledge \\
\href{https://zerobench.github.io/}{ZeroBench} & 100 & Challenging visual reasoning \\
\midrule
\textbf{Total (29 benchmarks)} & \textbf{11,221} & \\
\bottomrule
\end{tabular*}
\end{suppTable}

\clearpage
\subsection{Five-Benchmark Diagnostic Suite}
\label{sec:supp_five_benchmarks}

The diagnostic suite tests two steps of visual reasoning: finding the relevant detail and interpreting it in context. We select three benchmarks from MM400 for their emphasis on visual evidence. BabyVision covers fundamental visual reasoning, VLMsAreBiased tests whether image details override familiar priors, and BLINK tests core perceptual tasks such as correspondence and depth. We add VStarBench and MMVP to cover two complementary demands. VStarBench requires locating small cues in high-resolution scenes, and MMVP tests whether the model can distinguish inputs that look nearly identical.

The suite contains 1,679 questions (\cref{tab:supp_benchmarks,fig:benchmark_examples}). Macro-5 weights each benchmark equally.

\begin{suppTable}
\centering
\small
\setlength{\tabcolsep}{5pt}
\renewcommand{\arraystretch}{1.12}
\caption{\textbf{Five-benchmark diagnostic suite and evaluated snapshot sizes.}}
\label{tab:supp_benchmarks}
\begin{tabular*}{\linewidth}{@{\extracolsep{\fill}}lrlp{0.47\linewidth}@{}}
\toprule
Benchmark & Questions & Origin & Diagnostic focus \\
\midrule
BabyVision & 388 & MM400 & Discrimination, tracking, space, and patterns \\
VLMsAreBiased & 400 & MM400 & Local evidence that contradicts learned priors \\
BLINK & 400 & MM400 & Correspondence, depth, and spatial relations \\
VStarBench & 191 & Added & Search for small targets in high-resolution images \\
MMVP & 300 & Added & Distinguishing visually similar images \\
\midrule
\textbf{Total} & \textbf{1,679} & & \\
\bottomrule
\end{tabular*}
\end{suppTable}

\paragraph{BabyVision.}
BabyVision~\cite{chen2026babyvision} isolates elementary visual skills through tasks including pattern completion, tracking, and spatial judgments. Small changes to a mark, orientation, or position can change the answer, making it useful for testing which details survive image encoding.

\paragraph{VLMsAreBiased.}
VLMsAreBiased~\cite{vo2025biased} uses counterfactual variants of familiar visual patterns. A model may recognize the subject and still answer incorrectly unless it verifies the specific detail in the image, as in the stripe-counting example in \cref{fig:teaser}.

\paragraph{BLINK.}
BLINK~\cite{fu2024blink} casts classical computer vision tasks as questions over one or more images. It probes visual evidence such as point correspondence, depth order, and spatial relations. Answering can require keeping track of which part of which image a local feature belongs to.

\paragraph{VStarBench.}
VStarBench~\cite{wu2024vstar} places relevant objects or attributes in high-resolution scenes with many other details. It stresses locating the right region and then reading the local cue needed to answer.

\paragraph{MMVP.}
MMVP~\cite{tong2024eyes} is built around image pairs with similar representations but distinct visual attributes. Its questions test whether a model can distinguish the cue that changes the answer, such as an orientation or an object detail.

\clearpage
\begin{suppFigure}
\centering
\includegraphics[height=600bp]{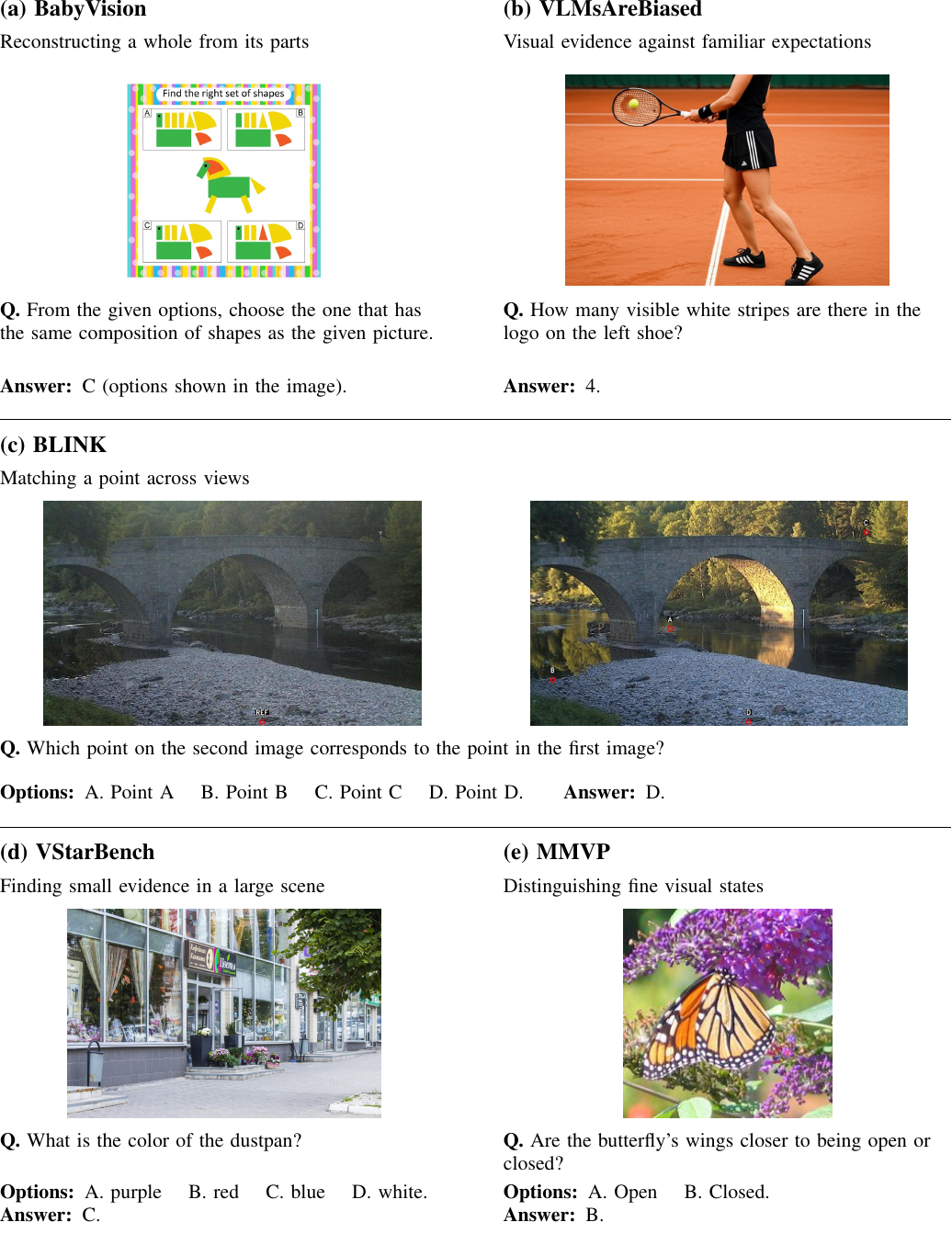}
\caption{\textbf{Five benchmarks, five visual demands.} Real image--question examples from our evaluation snapshots, with reference answers.}
\label{fig:benchmark_examples}
\end{suppFigure}
\clearpage
\section{BabyVision Subgroup Analysis}
\label{sec:three_statistics}

\subsection{Breadth Across Official BabyVision Task Types}
BabyVision's overall gain includes several distinct perceptual skills. \textbf{ResVision leads both baselines across all four official task types} (Fig.~\ref{fig:official_type_comparison}). The largest gain over VISTA is on fine-grained discrimination (+24.74 points), followed by pattern recognition (+18.30), spatial perception (+18.22), and visual tracking (+8.84).

\begin{suppFigure}
\includegraphics[width=0.58\textwidth]{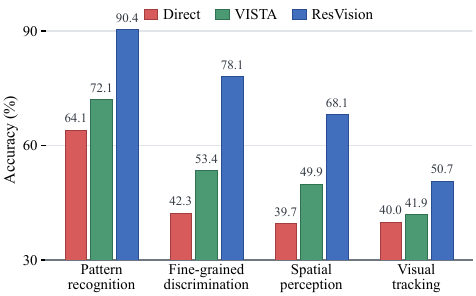}
\caption{\textbf{Breadth across official BabyVision types.} Accuracy averages four models and all three independent runs. The four types contain 51, 163, 91, and 83 questions, respectively.}
\label{fig:official_type_comparison}
\end{suppFigure}

\begin{suppTable}
\caption{\textbf{BabyVision gains by official task type.} Repeats and models are averaged within each question.}
\centering
\fontsize{8.5}{10.5}\selectfont
\setlength{\tabcolsep}{3.2pt}
\renewcommand{\arraystretch}{1.12}
\begin{tabular*}{0.88\linewidth}{@{\extracolsep{\fill}}lrr>{\columncolor{baselinegray}}rrr@{}}
\toprule
BabyVision type & $n$ & Direct & VISTA & ResVision & $\Delta$ \\
\midrule
Fine-grained discrim. & 163 & 42.33 & 53.37 & 78.12 & +24.74 \\
Pattern recognition & 51 & 64.05 & 72.06 & 90.36 & +18.30 \\
Spatial perception & 91 & 39.65 & 49.91 & 68.13 & +18.22 \\
Visual tracking & 83 & 39.96 & 41.87 & 50.70 & +8.84 \\
\bottomrule
\end{tabular*}

\end{suppTable}

\section{Failure Mode Taxonomy on the Five-Benchmark Suite}
\label{sec:error_taxonomy}

Accuracy shows how often each method answers correctly. To see where inference breaks down, we examine $4\times3\times1{,}679=20{,}148$ predictions per method for VISTA and ResVision on the five-benchmark suite (\cref{sec:supp_five_benchmarks}). A Kimi-K3 judge uses the question, reference answer, final response, and inspection trace to assign each incorrect prediction to one of six mutually exclusive categories. Error rates use all 20,148 predictions as the denominator.

\paragraph{Six failure modes.}
The categories distinguish what goes wrong after the model encounters a local cue, from assigning it to the wrong referent to tallying an incomplete set of instances.

\textbf{F1---referent binding} is an identity error: the model reads a local cue but attaches it to the wrong object, cell, image, or coordinate. In the D6 counting case (Appendix~\ref{sec:case-grid-address}), the crop makes the vertical marks legible but removes the row and column headers. The defining question is \emph{which entity or location does this evidence belong to?}

\textbf{F2---candidate-contrast omission} is a comparison error: the model accepts one plausible candidate without checking alternatives that could distinguish it. In the horse reconstruction case (Appendix~\ref{sec:case-reconstruction}), the extra yellow bar in A becomes decisive only when the candidate inventories are compared. The defining question is \emph{what distinguishes this candidate from the others?}

\textbf{F3---spatial/transform inconsistency} is a relational error: the model recognizes relevant local parts but combines them under an incorrect direction, rotation, topology, ordering, or cross-view correspondence. In the rotation case (Appendix~\ref{sec:case-rotation}), wedge colors can be read correctly while the requested transformation and answer option become mismatched. The defining question is \emph{how must these parts correspond after the stated relation or transformation?}

\textbf{F4---local-detail overinterpretation} is an interpretation error: the model identifies the correct entity and region but gives a local attribute the wrong meaning because the crop omits scale, pose, or object-level context. In the volleyball case (Appendix~\ref{sec:case-volleyball}), the arm-and-ball crop supports a plausible ``set'' interpretation, while the full pose supports ``bump.'' The defining question is \emph{what does this cue mean in the complete object or scene?}

\textbf{F5---coverage/count aggregation} collects failures to exhaustively enumerate repeated instances or reconcile partial counts. Its defining question is \emph{has the reasoning covered and reconciled every relevant instance?}

\textbf{F6---other errors} contains residual failures outside the preceding five classes.

\begin{suppFigure}
  \resizebox{0.92\linewidth}{!}{
\begingroup
\definecolor{errone}{HTML}{E76F51}
\definecolor{errtwo}{HTML}{F4A261}
\definecolor{errthree}{HTML}{5B8FF9}
\definecolor{errfour}{HTML}{6A994E}
\definecolor{errfive}{HTML}{7B61A8}
\definecolor{errsix}{HTML}{98A2B3}
\definecolor{correctbar}{HTML}{E9EDF3}
\definecolor{reportink}{HTML}{1B2738}
\begin{tikzpicture}[x=.118cm,y=1cm]
  \path[use as bounding box] (-18,-.48) rectangle (103,2.88);
  \tikzset{barlabel/.style={font=\bfseries\fontsize{9}{10.5}\selectfont,text=reportink,anchor=east},
    pct/.style={font=\bfseries\fontsize{9}{10.5}\selectfont,text=reportink,anchor=west},
    legend/.style={font=\fontsize{8.2}{9.6}\selectfont,text=reportink,anchor=west}}
  \node[barlabel] at (-2,2.10) {VISTA};
  \node[barlabel] at (-2,1.15) {ResVision};

  \fill[errone] (0,1.82) rectangle (3.0,2.38);
  \fill[errtwo] (3.0,1.82) rectangle (5.8,2.38);
  \fill[errthree] (5.8,1.82) rectangle (9.3,2.38);
  \fill[errfour] (9.3,1.82) rectangle (13.4,2.38);
  \fill[errfive] (13.4,1.82) rectangle (22.7,2.38);
  \fill[errsix] (22.7,1.82) rectangle (27.7,2.38);
  \fill[correctbar] (27.7,1.82) rectangle (100,2.38);
  \draw[reportink!35,line width=.5pt] (0,1.82) rectangle (100,2.38);
  \node[pct] at (101,2.10) {27.6\% error};

  \fill[errone] (0,.87) rectangle (2.1,1.43);
  \fill[errtwo] (2.1,.87) rectangle (4.1,1.43);
  \fill[errthree] (4.1,.87) rectangle (6.7,1.43);
  \fill[errfour] (6.7,.87) rectangle (10.3,1.43);
  \fill[errfive] (10.3,.87) rectangle (16.3,1.43);
  \fill[errsix] (16.3,.87) rectangle (19.9,1.43);
  \fill[correctbar] (19.9,.87) rectangle (100,1.43);
  \draw[reportink!35,line width=.5pt] (0,.87) rectangle (100,1.43);
  \node[pct] at (101,1.15) {19.8\% error};

  \foreach \x/\c/\t in {0/errone/F1 referent,18/errtwo/F2 contrast,36/errthree/F3 spatial,54/errfour/F4 interpretation,79/errfive/F5 coverage} {
    \fill[\c] (\x,-.08) rectangle +(2.2,.22);
    \node[legend] at (\x+2.9,.03) {\t};
  }
  \fill[errsix] (0,-.42) rectangle +(2.2,.22);
  \node[legend] at (2.9,-.31) {F6 other};
  \fill[correctbar] (18,-.42) rectangle +(2.2,.22);
  \draw[reportink!25,line width=.35pt] (18,-.42) rectangle +(2.2,.22);
  \node[legend] at (20.9,-.31) {correct};
\end{tikzpicture}
\endgroup
}
  \caption{\textbf{Error incidence over all predictions.} Each bar represents $20{,}148$ predictions. Colored segments denote the six failure classes and the pale remainder denotes correct predictions.}
  \label{fig:report_donut}
\end{suppFigure}

\begin{suppTable}
\centering
\small
\setlength{\tabcolsep}{5.4pt}
\renewcommand{\arraystretch}{1.16}
\caption{\textbf{Absolute incidence of system-level errors.} Percentages use the fixed 20,148 predictions per method.}
\label{tab:error_taxonomy_overall}
\begin{tabular}{l>{\columncolor{baselinegray}}rrrr}
\toprule
Error category & VISTA $n$ (\%) & ResVision $n$ (\%) & $\Delta n$ & Rel. reduction \\
\midrule
F1 Referent binding & $597\ (3.0)$ & $419\ (2.1)$ & $-178$ & $29.8\%$ \\
F2 Candidate contrast & $569\ (2.8)$ & $406\ (2.0)$ & $-163$ & $28.6\%$ \\
F3 Spatial consistency & $701\ (3.5)$ & $514\ (2.6)$ & $-187$ & $26.7\%$ \\
F4 Detail overinterpretation & $834\ (4.1)$ & $730\ (3.6)$ & $-104$ & $12.5\%$ \\
F5 Coverage/count aggregation & $1,867\ (9.3)$ & $1,201\ (6.0)$ & $-666$ & $35.7\%$ \\
F6 Other & $1,002\ (5.0)$ & $718\ (3.6)$ & $-284$ & $28.3\%$ \\
\midrule
\textbf{All errors} & $\mathbf{5,570\ (27.6)}$ & $\mathbf{3,988\ (19.8)}$ & $\mathbf{-1,582}$ & $\mathbf{28.4\%}$ \\
\bottomrule
\end{tabular}
\end{suppTable}

\begin{suppTable}
\centering
\small
\setlength{\tabcolsep}{3.8pt}
\renewcommand{\arraystretch}{1.16}
\caption{\textbf{Question-matched error transitions from VISTA to ResVision.} Fixed denotes VISTA-wrong/ResVision-correct predictions; regressed is the reverse. The category is assigned on the failing side.}
\label{tab:error_taxonomy_paired}
\begin{tabular}{l>{\columncolor{baselinegray}}rrrrrr}
\toprule
Error category & VISTA errors & Fixed & Regressed & Diff & Fix rate & Exact $p$ \\
\midrule
F1 Referent binding & $597$ & $301$ & $99$ & $+202$ & $50.4\%$ & $8.6\!\times\!10^{-25}$ \\
F2 Candidate contrast & $569$ & $313$ & $108$ & $+205$ & $55.0\%$ & $3.2\!\times\!10^{-24}$ \\
F3 Spatial consistency & $701$ & $335$ & $156$ & $+179$ & $47.8\%$ & $4.5\!\times\!10^{-16}$ \\
F4 Detail overinterpretation & $834$ & $325$ & $188$ & $+137$ & $39.0\%$ & $1.5\!\times\!10^{-9}$ \\
F5 Coverage/count aggregation & $1,867$ & $855$ & $223$ & $+632$ & $45.8\%$ & $1.1\!\times\!10^{-87}$ \\
F6 Other & $1,002$ & $389$ & $162$ & $+227$ & $38.8\%$ & $1.5\!\times\!10^{-22}$ \\
\midrule
\textbf{All errors} & $\mathbf{5,570}$ & $\mathbf{2,518}$ & $\mathbf{936}$ & $\mathbf{+1,582}$ & $\mathbf{45.2\%}$ & $<10^{-99}$ \\
\bottomrule
\end{tabular}
\end{suppTable}

\begin{suppTable}
\centering
\small
\setlength{\tabcolsep}{2.25pt}
\renewcommand{\arraystretch}{1.08}
\caption{\textbf{Complete error taxonomy by benchmark.} Entries are count (percentage of all predictions for that benchmark).}
\label{tab:error_taxonomy_benchmark}
\begin{tabular}{llrrrrrrr}
\toprule
Benchmark & Method & F1 Bind & F2 Contrast & F3 Spatial & F4 Detail & F5 Coverage & F6 Residual & All errors \\
\midrule
\multirow{2}{*}{BabyVision} & VISTA & \cellcolor{baselinegray}$380\ (8.2)$ & \cellcolor{baselinegray}$342\ (7.3)$ & \cellcolor{baselinegray}$407\ (8.7)$ & \cellcolor{baselinegray}$96\ (2.1)$ & \cellcolor{baselinegray}$564\ (12.1)$ & \cellcolor{baselinegray}$420\ (9.0)$ & \cellcolor{baselinegray}$2,209\ (47.4)$ \\
 & ResVision & $219\ (4.7)$ & $199\ (4.3)$ & $291\ (6.2)$ & $78\ (1.7)$ & $277\ (5.9)$ & $262\ (5.6)$ & $1,326\ (28.5)$ \\
\addlinespace[2pt]
\multirow{2}{*}{VLMsAreBiased} & VISTA & \cellcolor{baselinegray}$26\ (0.5)$ & \cellcolor{baselinegray}$56\ (1.2)$ & \cellcolor{baselinegray}$62\ (1.3)$ & \cellcolor{baselinegray}$215\ (4.5)$ & \cellcolor{baselinegray}$1,229\ (25.6)$ & \cellcolor{baselinegray}$330\ (6.9)$ & \cellcolor{baselinegray}$1,918\ (40.0)$ \\
 & ResVision & $6\ (0.1)$ & $49\ (1.0)$ & $42\ (0.9)$ & $216\ (4.5)$ & $863\ (18.0)$ & $272\ (5.7)$ & $1,448\ (30.2)$ \\
\addlinespace[2pt]
\multirow{2}{*}{BLINK} & VISTA & \cellcolor{baselinegray}$163\ (3.4)$ & \cellcolor{baselinegray}$166\ (3.5)$ & \cellcolor{baselinegray}$126\ (2.6)$ & \cellcolor{baselinegray}$124\ (2.6)$ & \cellcolor{baselinegray}$16\ (0.3)$ & \cellcolor{baselinegray}$214\ (4.5)$ & \cellcolor{baselinegray}$809\ (16.9)$ \\
 & ResVision & $158\ (3.3)$ & $152\ (3.2)$ & $102\ (2.1)$ & $108\ (2.2)$ & $9\ (0.2)$ & $152\ (3.2)$ & $681\ (14.2)$ \\
\addlinespace[2pt]
\multirow{2}{*}{VStarBench} & VISTA & \cellcolor{baselinegray}$26\ (1.1)$ & \cellcolor{baselinegray}$0\ (0.0)$ & \cellcolor{baselinegray}$56\ (2.4)$ & \cellcolor{baselinegray}$83\ (3.6)$ & \cellcolor{baselinegray}$5\ (0.2)$ & \cellcolor{baselinegray}$8\ (0.3)$ & \cellcolor{baselinegray}$178\ (7.8)$ \\
 & ResVision & $33\ (1.4)$ & $0\ (0.0)$ & $38\ (1.7)$ & $58\ (2.5)$ & $8\ (0.3)$ & $7\ (0.3)$ & $144\ (6.3)$ \\
\addlinespace[2pt]
\multirow{2}{*}{MMVP} & VISTA & \cellcolor{baselinegray}$2\ (0.1)$ & \cellcolor{baselinegray}$5\ (0.1)$ & \cellcolor{baselinegray}$50\ (1.4)$ & \cellcolor{baselinegray}$316\ (8.8)$ & \cellcolor{baselinegray}$53\ (1.5)$ & \cellcolor{baselinegray}$30\ (0.8)$ & \cellcolor{baselinegray}$456\ (12.7)$ \\
 & ResVision & $3\ (0.1)$ & $6\ (0.2)$ & $41\ (1.1)$ & $270\ (7.5)$ & $44\ (1.2)$ & $25\ (0.7)$ & $389\ (10.8)$ \\
\bottomrule
\end{tabular}
\end{suppTable}

ResVision reduces errors from 5,570 (27.6\%) to 3,988 (19.8\%), a relative reduction of 28.4\%. Coverage and count aggregation shows the largest relative reduction (35.7\%). We also compare paired predictions to see which questions change from correct to incorrect or vice versa. ResVision corrects 38.8--55.0\% of VISTA errors within each class, and corrections outnumber regressions in all six. In total, 2,518 predictions are corrected and 936 regress, a net gain of 1,582 ($p<10^{-99}$, exact two-sided McNemar test). Table~\ref{tab:error_taxonomy_benchmark} reports the per-benchmark distribution, and Appendix~\ref{sec:supp-qualitative} traces individual examples.

\section{Representative Failure Cases}
\label{sec:supp-qualitative}

The taxonomy identifies five recurring ways a model can misuse local evidence (Appendix~\ref{sec:error_taxonomy}). Each case below follows one such error through the question, the visual evidence, and the three methods' answers. The examples show what the larger scene contributes, from the row and column that identify a cell to the full pose that gives a gesture its meaning.

\begin{suppTable}
  \centering
  \caption{\textbf{Severity-ordered qualitative cases.} Entries are exact-match successes out of three runs.}
  \label{tab:qual-stability}
  \small
  \setlength{\tabcolsep}{4pt}
  \renewcommand{\arraystretch}{1.12}
  \begin{tabular}{clllccc}
    \toprule
    Case & Category & Benchmark / ID & Global constraint & Direct & VISTA & ResVision \\
    \midrule
    A & F1 Referent binding & VLMsAreBiased / 62280 & cell address & 1/3 & 1/3 & \textbf{3/3} \\
    B & F3 Spatial consistency & BabyVision / 510 & row--transform mapping & 0/3 & 1/3 & \textbf{3/3} \\
    C & F5 Coverage aggregation & BabyVision / 4552 & complete lattice & 0/3 & 0/3 & \textbf{3/3} \\
    D & F2 Candidate contrast & BabyVision / 2353 & shared inventory & 1/3 & 0/3 & \textbf{3/3} \\
    E & F4 Detail interpretation & MMVP / 90100 & whole-body pose & \textbf{3/3} & 0/3 & \textbf{3/3} \\
    \bottomrule
  \end{tabular}
\end{suppTable}

\subsection{Case A (F1): Bind the Detail to the Right Referent}
\label{sec:case-grid-address}

The question asks how many vertical marks appear in cell D6.  Enlargement makes the marks easier to distinguish, while the full table supplies their row and column address. A correct local count answers the question only if it is bound to D6.

\begin{itemize}
  \setlength{\itemsep}{1pt}
  \item \textbf{Direct} answers $4,3,4$.  It is correct once, but the count changes across runs because the small marks are difficult to read reliably in the full table.
  \item \textbf{VISTA} answers $2,4,3$.  Its crop makes the marks larger, but removes the row and column headers.  The model can therefore read a nearby cell clearly while losing confidence about whether that cell is D6.
  \item \textbf{ResVision} answers $3,3,3$.  It first locates column D and row 6 in the full table, then enlarges only their intersection and counts the three marks there.
\end{itemize}
\begin{suppFigure}
  \centering
  \includegraphics[width=0.98\linewidth]{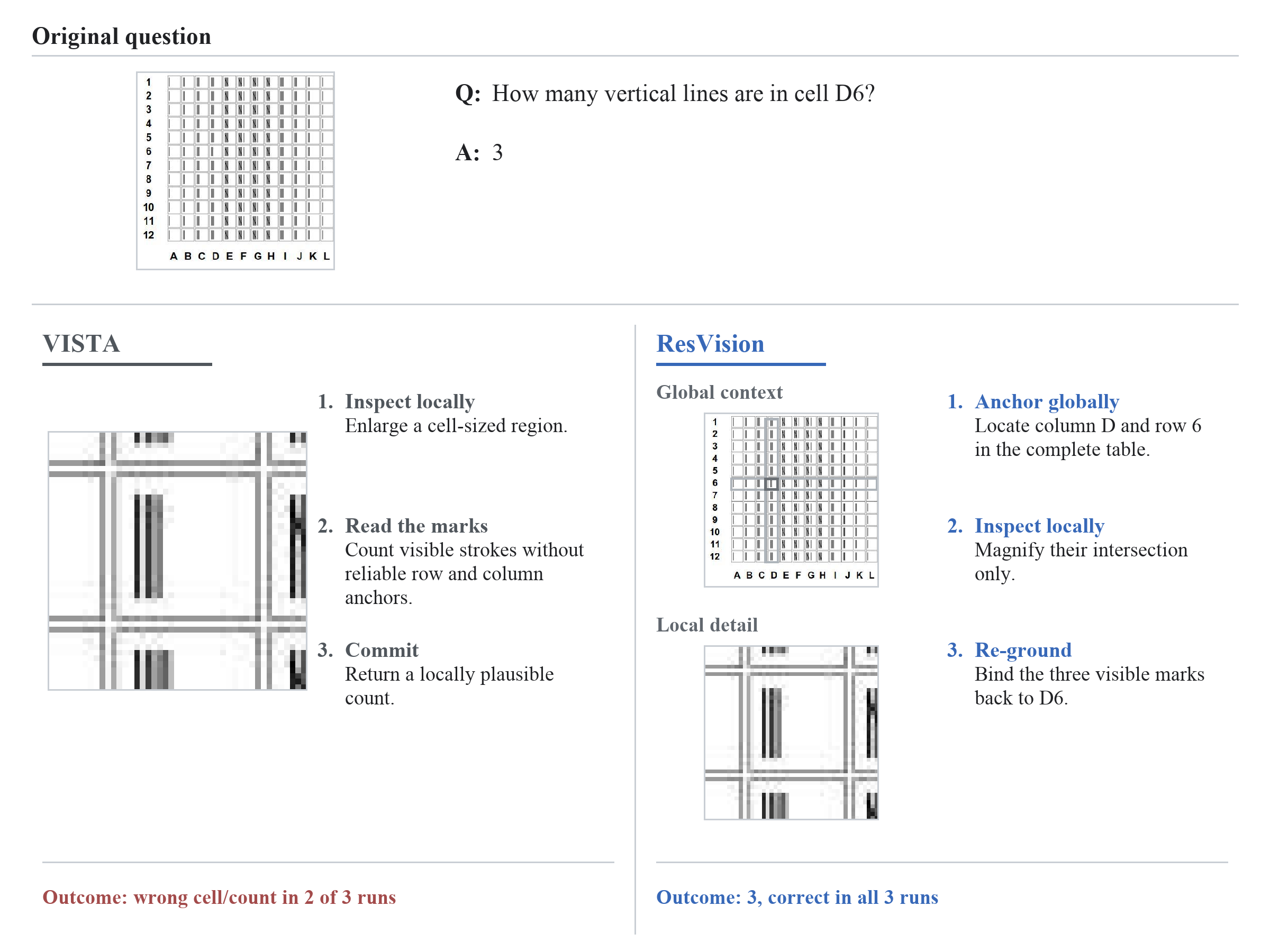}
  \caption{\textbf{F1---referent binding.} The crop shows the marks and the full table identifies cell D6.}
  \label{fig:case-grid-address}
\end{suppFigure}

\clearpage
\subsection{Case B (F3): Preserve the Spatial Contract}
\label{sec:case-rotation}

The question asks for three different rotations and expects an ordered triple.  Each local match must therefore stay linked to its source row, requested angle, and final option number.

\begin{itemize}
  \setlength{\itemsep}{1pt}
  \item \textbf{Direct} answers $(6,4,5)$, $(6,4,5)$, and $(5,1,3)$.  It can recognize some rotated patterns, but does not keep all three row-specific correspondences correct at once.
  \item \textbf{VISTA} answers $(4,6,2)$, $(6,2,3)$, and $(6,4,2)$.  Enlarged wedges clarify the local color order, yet the row-to-angle-to-option bookkeeping is lost in two runs.
  \item \textbf{ResVision} answers $(6,4,2)$ in all three runs.  It uses local views to resolve wedge order while the full layout keeps row 1, row 2, and row 3 tied to the right transformations and options.
\end{itemize}
\begin{suppFigure}
  \centering
  \includegraphics[width=0.98\linewidth]{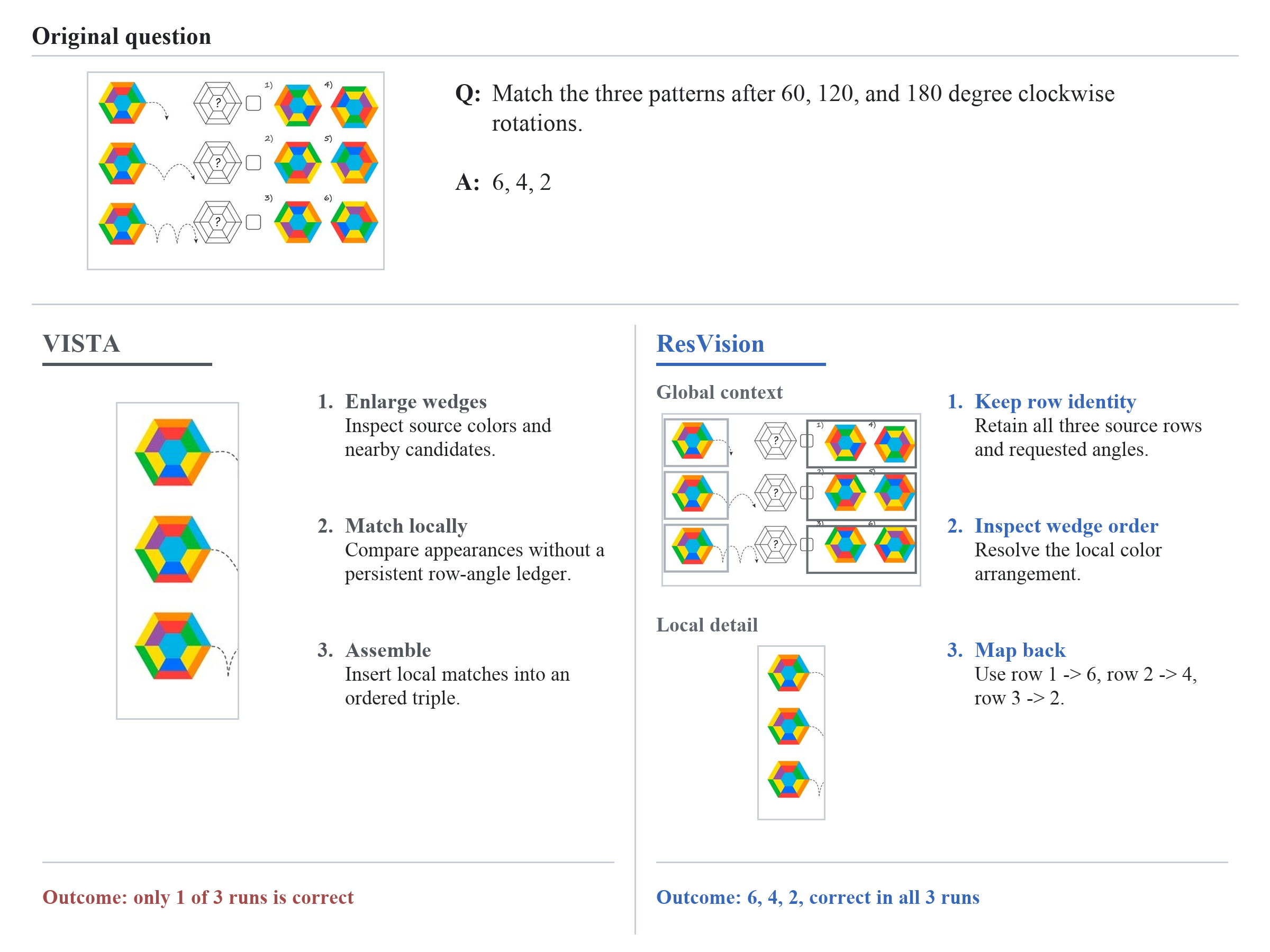}
  \caption{\textbf{F3---spatial consistency.} Source rows, rotation angles, and matched options.}
  \label{fig:case-rotation}
\end{suppFigure}

\clearpage
\subsection{Case C (F5): Make Coverage Auditable}
\label{sec:case-grid-count}

The question asks for the total number of dark cells in a $10\times11$ grid.  Individual cells are easy to identify, but the final count is correct only if every position is visited once, with no omissions or duplicates.

\begin{itemize}
  \setlength{\itemsep}{1pt}
  \item \textbf{Direct} answers $37,39,46$.  All three totals are too small, which is consistent with losing cells while reading a dense grid at its original scale.
  \item \textbf{VISTA} answers $59,60,52$.  Zooming makes a local group of cells easier to read, but the totals move above and below the truth because there is no explicit record of which positions have already been counted.
  \item \textbf{ResVision} answers $54,54,54$.  It treats the full lattice as a ledger, records the row totals $2,3,5,7,3,7,10,8,4,5,0$, and checks that every cell contributes exactly once before summing them.
\end{itemize}
\begin{suppFigure}
  \centering
  \includegraphics[width=0.98\linewidth]{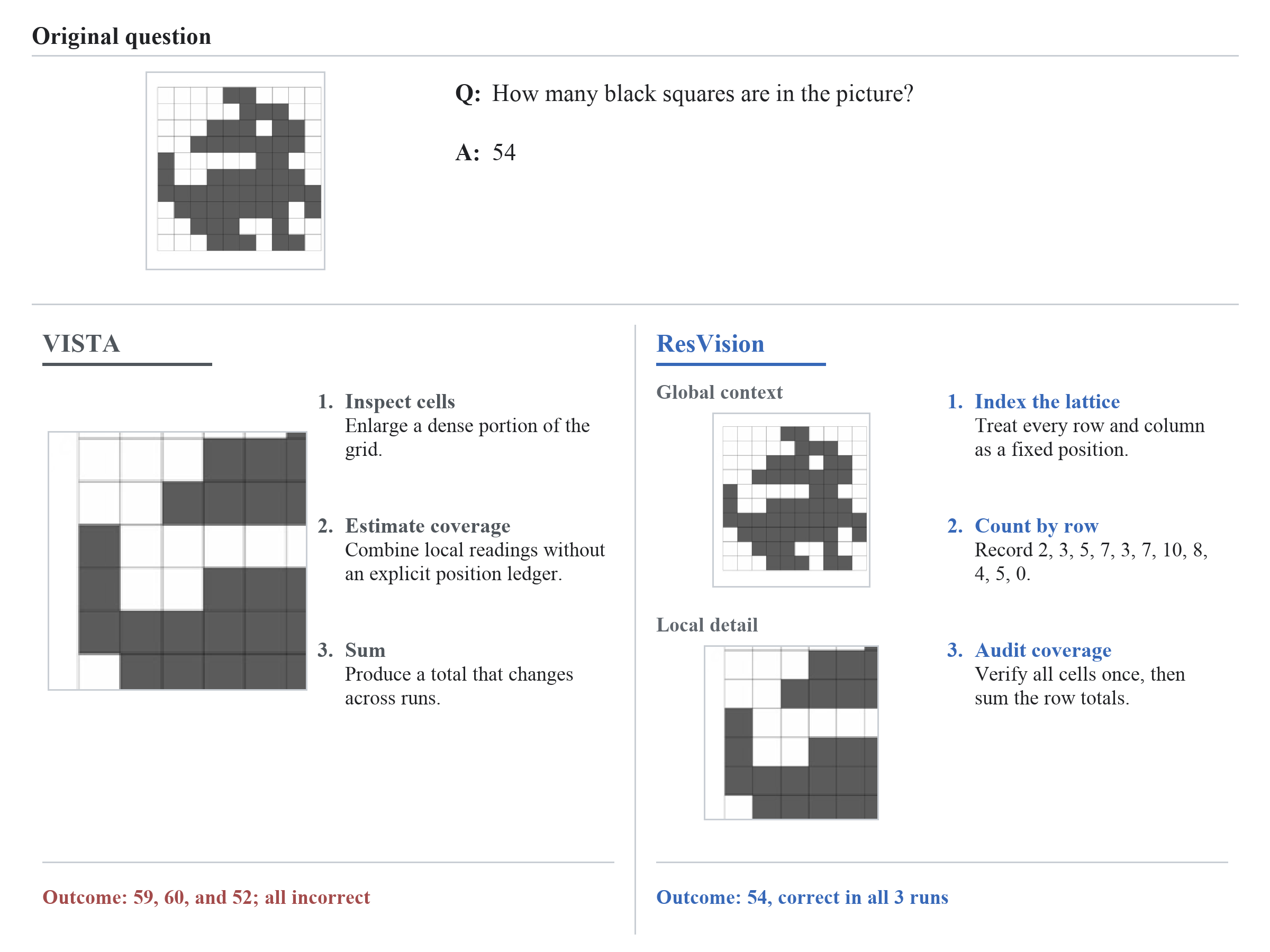}
  \caption{\textbf{F5---coverage aggregation.} Row-by-row count of dark cells.}
  \label{fig:case-grid-count}
\end{suppFigure}

\clearpage
\subsection{Case D (F2): Compare the Competing Hypotheses}
\label{sec:case-reconstruction}

The question asks which option contains exactly the same pieces as the target horse.  Options A and C look similar, so the decision depends on comparing their complete piece inventories rather than describing the target alone.

\begin{itemize}
  \setlength{\itemsep}{1pt}
  \item \textbf{Direct} answers A, A, and C.  It notices the overall resemblance, but only one run distinguishes the candidates correctly.
  \item \textbf{VISTA} answers A in all three runs.  Its close view gives a detailed description of the target horse, yet does not place A and C in the same comparison; the extra yellow bar in A is therefore overlooked.
  \item \textbf{ResVision} answers C in all three runs.  It compares the target, A, and C together, checks the piece inventory item by item, and rejects A after finding its extra yellow bar.
\end{itemize}
\begin{suppFigure}
  \centering
  \includegraphics[width=0.98\linewidth]{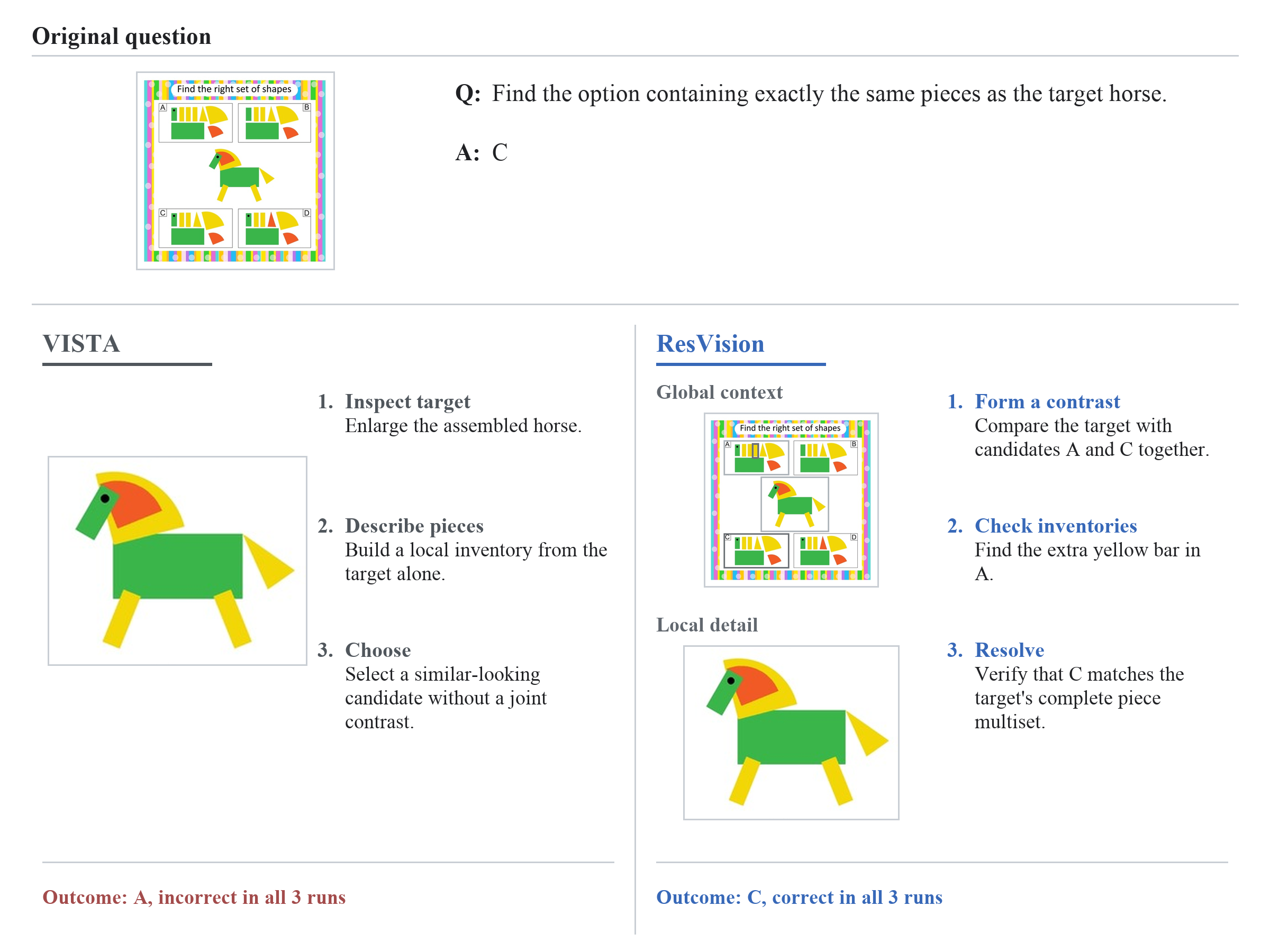}
  \caption{\textbf{F2---candidate contrast.} Piece inventories of the target and options A and C.}
  \label{fig:case-reconstruction}
\end{suppFigure}

\clearpage
\subsection{Case E (F4): Interpret the Detail in the Whole Pose}
\label{sec:case-volleyball}

The question asks whether the player is more likely to bump or set the volleyball.  A close view of the arms and ball can suggest either action; the whole-body pose is needed to decide which interpretation is physically consistent.

\begin{itemize}
  \setlength{\itemsep}{1pt}
  \item \textbf{Direct} answers A (bump) in all three runs.  In this example the full image already preserves the decisive cue: the player is leaning forward with joined forearms held low.
  \item \textbf{VISTA} answers B (set) in all three runs.  The crop emphasizes a bent arm and a high ball, which supports a plausible local story about setting, but hides the rest of the posture that contradicts it.
  \item \textbf{ResVision} answers A (bump) in all three runs.  It inspects the arm-and-ball region while retaining the full pose, then checks that the local cue agrees with the joined forearm platform and forward body geometry.
\end{itemize}
\begin{suppFigure}
  \centering
  \includegraphics[width=0.98\linewidth]{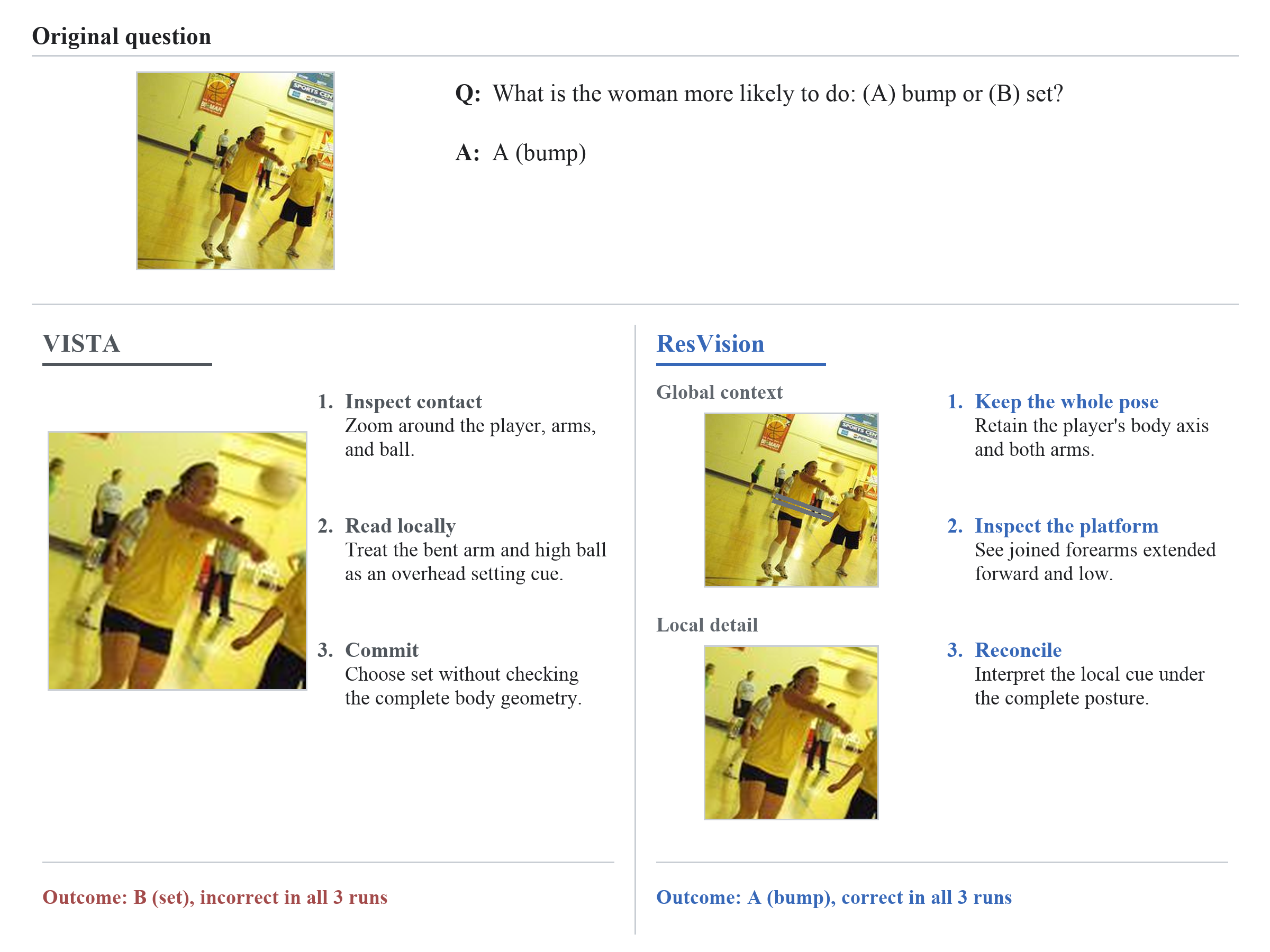}
  \caption{\textbf{F4---detail interpretation.} A close-up of the arms alongside the player's full pose.}
  \label{fig:case-volleyball}
\end{suppFigure}

\stopcontents[appendix]
\fi

\end{document}